\documentclass[3p,twocolumn]{elsarticle}

\usepackage{lineno,hyperref}
\modulolinenumbers[5]

\usepackage{graphicx}
\usepackage{booktabs}
\usepackage{multirow}
\usepackage{amsmath}
\usepackage{amsfonts,amssymb}  
\usepackage{subfigure}
\usepackage{algorithm}
\usepackage{algpseudocode}
\usepackage{mathtools}
\usepackage{graphicx,subfigure}
\usepackage{color}
\usepackage{epstopdf}

\graphicspath{{figures/}{qaudEXfigure/}{ThreeObs/}}

\newcommand{\argmin}{\operatornamewithlimits{argmin}}
\newcommand{\absmin}{\operatornamewithlimits{absmin}}

\newtheorem{thm}{Theorem}
\newtheorem{cor}[thm]{Corollary}

\newcommand{\RE}[1]{{\color{blue}#1}} 

\begin{document}
\allowdisplaybreaks

\begin{frontmatter}

\title{Topology-Guided Path Integral Approach for Stochastic Optimal Control in Cluttered Environment} 

\author{Jung-Su Ha, Soon-Seo Park}
\author{Han-Lim Choi\corref{mycorrespondingauthor}}
\address{KAIST, 291 Daehak-ro, Yuseong, Deajeon 34141, Republic of Korea.}
\cortext[mycorrespondingauthor]{Corresponding author}
\ead[mycorrespondingauthor]{hanlimc@kaist.ac.kr}

\begin{keyword}                           
Stochastic Optimal Control\sep Topological Motion Planning\sep Linearly-Solvable Optimal Control\sep Multi-modality              
\end{keyword}                             

\begin{abstract}
This paper addresses planning and control of robot motion under uncertainty that is formulated as a continuous-time, continuous-space stochastic optimal control problem, by developing a topology-guided path integral control method. The path integral control framework, which forms the backbone of the proposed method, re-writes the Hamilton-Jacobi-Bellman equation as a statistical inference problem; the resulting inference problem is solved by a sampling procedure that computes the distribution of controlled trajectories around the trajectory by the passive dynamics. For motion control of robots in a highly cluttered environment, however, this sampling can easily be trapped in a local minimum unless the sample size is very large, since the global optimality of local minima depends on the degree of uncertainty. Thus, a homology-embedded sampling-based planner that identifies many (potentially) local-minimum trajectories in different homology classes is developed to aid the sampling process. In combination with a receding-horizon fashion of the optimal control the proposed method produces a dynamically feasible and collision-free motion plans without being trapped in a local minimum. Numerical examples on a synthetic toy problem and on quadrotor control in a complex obstacle field demonstrate the validity of the proposed method.

\end{abstract}

\end{frontmatter}

\section{Introduction}

Computing the optimal policy for a system driven by some uncertain disturbance, which is called a \textit{stochastic optimal control problem}, is one of the most important problems in planning/control of robotic platforms in a cluttered environment.
In a discrete-time/discrete-state and control space setting, the problem is formulated as a Markov decision process (MDP) and solved through the dynamic programming procedure, e.g. value iteration or policy iteration.
The problem in a continuous setting, which is of the primary interest of this work, can be solved in a similar manner if transformed into a discretized version; however, this discretization approach is not scalable for a high-dimensional state space.
Alternatively, an optimality condition for the continuous problem itself can be derived and utilized. It is well known that the optimality condition results in a nonlinear partial differential equation (PDE), called the Hamilton-Jacobi-Bellman equation; but, solving a nonlinear PDE is intractable in most robotic control applications.

Fortunately, there is a class of stochastic optimal control problem, called linearly-solvable optimal control (LSOC)~\cite{todorov2009efficient}, for which the HJB equation can be solved in a more efficient way with appropriate reformulation.
For an LSOC problem, the notion of a \textit{desirability} function, which is effectively an exponential value function, is introduced in order to transcribe the original nonlinear HJB equation on the value function into a linear PDE on the desirability function. In addition, it has been found that the Feynman-Kac formula allows the solution of such linear PDE to be expressed as an expectation of some path integral.
As a result, the stochastic optimal control problem is transformed into an estimation problem, which can be solved by sampling a set of stochastic paths and then evaluating their expectation.
This aforementioned procedure to solve a LSOC is referred to as path integral (PI) control~\cite{kappen2005path}.
For more interesting views and different derivations of PI control, we would refer the reader to \cite{theodorou2015nonlinear} and references therein.

Advanced estimation techniques, such as importance sampling, can be applied to effectively solve the aforementioned transformed  problem of a LSOC.
In \cite{theodorou2010generalized, theodorou2010reinforcement}, the control policy is parameterized and then estimated using an importance sampling technique on the basis of the path integral formula.
In \cite{thijssen2015path}, path-integral formula is utilized to construct a state-dependent feedback controller and theoretical analysis on how sampling strategies affect the estimation results is presented.
In \cite{kappen2015adaptive}, the cross entropy method was applied to build an efficient importance sampler that reduces estimation variance.
In \cite{arslan2014information}, the rapidly-exploring random tree (RRT) algorithm was used to help the importance sampler to pick valuable samples.

This work addresses a continuous LSOC problem, especially in a complex configuration space with obstacles, in the path integral control framework.
This type of problem may have many local optima, since  the state space is often highly non-convex due to obstacle regions.
Thus, a sampler for PI control needs to be able to generate samples diverse and spread enough in order not to be trapped into a local minimum; however, it is not particularly easy for many conventional sampling schemes to generate samples very far from most of other samples.
To tackle this issue, the approach in this work, therefore, (i) first specifies all possible local minima caused by obstacles for deterministic approximation of the original problem and then (ii) generates samples around all these local minima taking them as \textit{reference trajectories}.
If the global minimum of the original problem is near one of these \textit{references}, this way eventually results in finding the global optimal solution.

Specifically in the context of motion planning in a cluttered environment, each local minimum can be associated with a different topological class; thus, a motion planner that can produce a optimized motion trajectory for every different topological class is required to support the above two-step process. There have been some attempts to build a topology-embedded path planner (although not in the context of stochastic control).
One of the most general topological concept is homology;
two trajectories are in the same homology class, if the boundary formed by one trajectory together with the (opposite directional) other one does not contain any obstacles.
There have been some attempts to embed the concept of homology in motion planning algorithms.
Bhattacharya et al. have proposed the concept of $H$-signature to distinguish different homology classes of trajectories and incorporated it into a graph-search algorithm to find the optimal trajectories in various homology classes for 2--3 dimensional \cite{bhattacharya2012topological} and higher dimensional \cite{bhattacharya2013invariants} configuration space;
they have augmented the configuration space by $H$-signature and performed A* algorithm on $H$-signature augmented graph.
$H$-signature has also been adopted in \cite{rosmann2017integrated} to enumerate all different homology classes of trajectories with a Voronoi diagram and to optimize each trajectory with a local optimizer.
In \cite{pokorny2016topologicalijrr,pokorny2016topological}, Pokorny et al. have proposed the algebraic topological approach to automatically distinguish different homology classes of trajectories without explicit information about obstacle positions by utilizing filtration of simplicial complexes.
Also, a topological task projection is proposed in \cite{pokorny2016high} to represent topological features of high-dimensional trajectories by $H$-signature in 2-dimensional projected space.

This concept of $H$-signature is valid, but it is known to be difficult for the graph search algorithm to handle high-dimensional state space and system dynamics.
In the motion planning literature, sampling-based algorithms have widely been studied in order to cope with such difficulties and made a lot of successes theoretically and practically \cite{lavalle2006planning}. 
Especially, Karaman and Frazzoli have proposed the incremental sampling-based algorithm, namely the Rapidly-exploring Random Tree star (RRT*) \cite{karaman2011sampling}, and more recently, Janson et al. have proposed the Fast Marching Tree star (FMT*) algorithm \cite{janson2015fast} which utilizes batch process; both algorithms guarantee probabilistic completeness and asymptotic optimality.
They have naturally extended to the planning problem with high-dimensional space and system dynamics \cite{karaman2010optimal, webb2013kinodynamic, ha2013successive, allen2015toward}.
Very few attempts, however, have been made at adopting sampling-based algorithm to topological motion planning problem whose configuration space is augmented by topological signature.
The Probabilistic Roadmap-based and the RRT-based approaches have been proposed, which are capable of generating paths corresponding to as many homotopic groups as possible~\cite{kala2016homotopic,hernandez2015comparison}; the objective of that work is not to find the optimal trajectory, but to identify many trajectories in different homotopy classes.
Only very recent research, Winding-Augmented RRT* (WA-RRT*) \cite{pokorny2016high} and Rapidly-exploring Random Homology-embedded Tree star (RRHT*)~\cite{ha2016topology}, similarly extent RRT* algorithm to topological optimal motion planning; they attempt to find the optimal trajectories in different homotopy classes.
WA-RRT* conducts an \textit{additional} $H$-signature sampling step; in order to create a new node, it samples a value of $H$-signature as well as its configuration coordinate.
RRHT*, on the other hand, expands a graph in the state space using the rapidly-exploring random graph (RRG) \cite{karaman2011sampling} algorithm and projects an associated tree onto the $H$-signature augmented space.

This paper presents an algorithm, termed Path-Integral with Fast Marching Homology-embedded Tree star (PI-FMHT*), that consists of a homology-embedded optimal motion planner to identify the local minima of deterministic approximation to the original problem and an importance sampler that solves a transformed estimation problem of the original LSOC.
Combined with a receding-horizon scheme for plan \& execution of the stochastic optimal solution, the proposed method can produces the globally optimal, dynamically feasible collision-free trajectory for stochastic systems.
While a brief idea of topology-guided path integral control methodology has been introduced in the authors' preliminary work~\cite{ha2016topology}, this paper proposes a much more efficient topological motion planner based on FMT* algorithm, includes much detailed description and comparison of the methodology, as well as more diverse and extensive numerical case studies.

\section{Linearly-Solvable Stochastic Optimal Control}
\subsection{Problem Description}
Suppose $\mathbf{x}\in \mathbb{R}^n$ and $\mathbf{u}\in \mathbb{R}^m$ are a state and control vector, respectively, $\mathbf{w}$ is an $m$-dimensional Wiener process.
Consider the stochastic dynamics of which deterministic drift term is affine in control input:
\begin{equation}
d\mathbf{x} =  \mathbf{f}(\mathbf{x})dt+ G(\mathbf{x})\mathbf{u}dt + B(\mathbf{x})d\mathbf{w} \label{eq:dyn_control}
\end{equation}
where $\mathbf{f}:\mathbb{R}^n\rightarrow \mathbb{R}^n$ is the passive dynamics and $G:\mathbb{R}^n\rightarrow \mathbb{R}^{n\times m}$ is control transition matrix and $B:\mathbb{R}^n\rightarrow \mathbb{R}^{n\times m}$ is the diffusion matrix function.
In this work, the state is assumed to be partitioned as $\mathbf{x}=[\mathbf{x}_m^T~\mathbf{x}_c^T]^T$ and then other terms are partitioned as $\mathbf{f}(\mathbf{x}) = [\mathbf{f}_m(\mathbf{x})^T~\mathbf{f}_c(\mathbf{x})^T]^T$, $G(\mathbf{x}) = [\mathbf{0}_{(n-m)\times m}^T~G_c(\mathbf{x})^T]^T$ and $B(\mathbf{x}) = [\mathbf{0}_{(n-m)\times m}^T~B_c(\mathbf{x})^T]^T$.
It is also assumed that $G_c:\mathbb{R}^n\rightarrow \mathbb{R}^{m\times m}$ and $B_c:\mathbb{R}^n\rightarrow \mathbb{R}^{m\times m}$ are invertible.

The objective of the problem is to find a control policy which achieves the goal region while avoiding collision with other boundaries (e.g. obstacles) and also minimizes the control effort and/or the state cost.
We formulate the problem as a finite-horizon stochastic optimal control problem with a fixed final time $t_f$.
Let a function $q: \mathbb{R}^n \rightarrow \mathbb{\bar{R}}$ and $\phi:\mathbb{R}^n \rightarrow \mathbb{\bar{R}}$ be an instantaneous state cost rate and a final cost function, respectively, where $\mathbb{\bar{R}}$ denotes the extended real number line $\mathbb{R}\cup\{-\infty,+\infty\}$.
For given control policy $\pi:\mathbb{R}\times\mathbb{R}^n\rightarrow\mathbb{R}^m$, the cost functional which we want to minimize is defined as:	
\begin{equation}
J^\mathbf{\pi}(t,\mathbf{x}) = E\left[\phi(\mathbf{x}(t_f))+\int^{t_f}_t q(\mathbf{x})+\frac{1}{2}\mathbf{u}^TR(\mathbf{x})\mathbf{u}d\tau\right],
\end{equation}
where $\mathbf{x}(t)$ is a solution of \eqref{eq:dyn_control} with $\mathbf{u}(t)=\pi(t,\mathbf{x}(t))$.
The instantaneous state cost rate, $q(\cdot)$, encodes a penalty for collision with an obstacle or preference of certain states, the final cost function, $\phi(\cdot)$, penalizes a distance of final state from the goal, and $R$ is a matrix for the control penalty.

\subsection{Path Integral Control}
\RE{The optimal cost-to-go function is defined as:
\begin{equation}
v(t,\mathbf{x}) \equiv \inf_\pi J^\pi(t,\mathbf{x}),
\end{equation}
and the associated Hamilton-Jacobi-Bellman (HJB) equation is given by:
\begin{align}
-v_t =\min_\mathbf{u}(q+\frac{1}{2}\mathbf{u}^TR\mathbf{u}&+(f+G\mathbf{u})^Tv_\mathbf{x} \nonumber\\
&~~~~+\frac{1}{2}\text{tr}(BB^Tv_{\mathbf{xx}})), \label{eq:HJB2}
\end{align}
with $v(t_f,\mathbf{x})=\phi(\mathbf{x})$ by definition, where subscript notations are used to represent partial derivatives, i.e., $v_t=\frac{\partial v}{\partial t},~v_\mathbf{x}=\frac{\partial v}{\partial\mathbf{x}}$ and $v_\mathbf{xx}=\frac{\partial^2 v}{\partial\mathbf{x}^2}$.}
From the HJB equation, the optimal control law is obtained analytically as:
\begin{equation}
\mathbf{u}^*(t,\mathbf{x}) = -R^{-1}(\mathbf{x})G^T(\mathbf{x})v_\mathbf{x}(t,\mathbf{x}).\label{eq:opt_cont}
\end{equation}
Substituting this optimal control law to (\ref{eq:HJB2}) yields the second order nonlinear partial differential equation (PDE):
\begin{align}
-v_t = q(\mathbf{x})+v_\mathbf{x}^Tf(\mathbf{x})&-\frac{1}{2}v_\mathbf{x}^TG(\mathbf{x})R^{-1}(\mathbf{x})G^T(\mathbf{x})v_\mathbf{x} \nonumber\\
&+\frac{1}{2}\text{tr}(v_{\mathbf{xx}}B(\mathbf{x})B^T(\mathbf{x})).\label{eq:non_PDE}
\end{align}

Due to its nonlinearity, solving the above PDE is intractable.
The nonlinearity can be removed by introducing the \textit{desirability function}:
\begin{equation}
\psi(t,\mathbf{x}) = \exp(-\frac{1}{\lambda}v(t,\mathbf{x})),
\end{equation}
where a scalar, $\lambda$ comes from the relation,
\begin{equation}
\lambda G(\mathbf{x})R^{-1}(\mathbf{x})G^T(\mathbf{x}) = B(\mathbf{x})B^T(\mathbf{x}).
\end{equation}
This restriction means that the control and noise affect the dynamics on the same subspace and in the same direction and the control cost is reversely related to the noise scale \cite{kappen2005path,todorov2009efficient}.
Roughly speaking, with the above restriction the control is more expensive for the direction that the noise is smaller.
Rewriting the PDE in (\ref{eq:non_PDE}) with respect to $\psi(\mathbf{x})$ induces the second order linear PDE as:
\begin{equation}
-\psi_t = -\frac{1}{\lambda}q(\mathbf{x})\psi+f^T(\mathbf{x})\psi_\mathbf{x}+\frac{1}{2}\text{tr}(\psi_{\mathbf{xx}} B(\mathbf{x})B^T(\mathbf{x})), \label{eq:lin_PDE}
\end{equation}
where the final condition is given by:
\begin{equation}
\psi(t_f,\mathbf{x}) = \exp(-\frac{1}{\lambda}\phi(\mathbf{x})). \label{eq:lin_PDE_bnd}
\end{equation}

The problem in (\ref{eq:lin_PDE}) and (\ref{eq:lin_PDE_bnd}) is called the Cauchy problem \cite{karatzas2012brownian} and its solution can be represented probabilistically by the Feynman-Kac formula.
Following corollary is directly modified from Proposition 5.7.6 in \cite{karatzas2012brownian}.
\begin{thm}[Feynman-Kac]
	Let $\mathbf{x}(t)$ be a solution of
	\begin{equation}
	d\mathbf{x} = f(\mathbf{x})dt + B(\mathbf{x})d\mathbf{w}^{(0)}. \label{eq:dyn_passive}
	\end{equation}
	Suppose $\psi(t,\mathbf{x})$ is continuous and satisfies the  Cauchy problem (\ref{eq:lin_PDE}) and (\ref{eq:lin_PDE_bnd}).
	Then, $\psi(t,\mathbf{x})$ admits the stochastic representation:
	\begin{align}
	&\psi(t,\mathbf{x}) =\nonumber\\
	& E_P\left[\exp\left(-\frac{1}{\lambda}\left(\phi(\mathbf{x}(t_f))+\int^{t_f}_tq(\mathbf{x}(\tau))d\tau\right)\right)\right], \label{eq:desir_fun}
	\end{align}
	where the expectation $E_P[\cdot]$ is taken over all trajectories $\mathbf{x}(t),~t\in[0,t_f]$.
\end{thm}

The optimal control (\ref{eq:opt_cont}) is written with respect to $\psi$ as:
\begin{align}
\mathbf{u}^*(t,\mathbf{x}) &= \lambda R^{-1}(\mathbf{x})G^T(\mathbf{x})\frac{\psi_\mathbf{x}(t,\mathbf{x})}{\psi(t,\mathbf{x})}\nonumber\\
& = \lambda R^{-1}(\mathbf{x})G_c^T(\mathbf{x})\frac{\psi_\mathbf{x_c}(t,\mathbf{x})}{\psi(t,\mathbf{x})}.\label{eq:opt_cont2}
\end{align}
Equation (\ref{eq:desir_fun}) can be expressed as
\begin{equation}
\psi(t,\mathbf{x}) = \int W(\vec{\mathbf{x}}) P(\vec{\mathbf{x}})d\vec{\mathbf{x}}, \label{eq:est_cost}
\end{equation}
where $W(\vec{\mathbf{x}}) = \exp\left(-\frac{1}{\lambda}\left(\phi(\mathbf{x}(t_f))+\int^{t_f}_tq(\mathbf{x(\tau)})d\tau\right)\right)$ and $\vec{\mathbf{x}}$ and $P(\vec{\mathbf{x}})$ represent trajectories and its probability measure, respectively.
From the path integral formulation~\cite{kappen2005path}, the probability measure of trajectory is given by:
\begin{equation}
P(\vec{\mathbf{x}}) = c\lim_{dt\rightarrow 0}\exp\left(-\frac{1}{2\lambda}\sum_{j=1}^{N} \left[ \left\| \mu(\mathbf{x}_j) \right\|^2 _{\Sigma_c^{-1}(\mathbf{x}(t_j))}\right] dt\right),
\end{equation}
where $t_1 = t,~t_N = t_f$, $\Sigma_c(\mathbf{x}) = G_c(\mathbf{x})R^{-1}(\mathbf{x})G_c^T(\mathbf{x}) =B_c(\mathbf{x})B_c^T(\mathbf{x})/\lambda$ and $\mu(\mathbf{x}_j)\equiv \frac{\mathbf{x}_c(t_j+dt)-\mathbf{x}_c(t_j)}{dt}-\mathbf{f}_c(\mathbf{x}(t_j))$ and $c$ is a normalization constant for $\int dP(\vec{\mathbf{x}})=1$.
Partial derivative of $P$ is given by:
\begin{equation}
\frac{\partial}{\partial \mathbf{x}_c(t_1)}P(\vec{\mathbf{x}}) = \frac{1}{\lambda}\mu^T(\mathbf{x}_1)\Sigma_c^{-1}(\mathbf{x}(t_1))P(\vec{\mathbf{x}}),
\end{equation}
which yields
\begin{align}
\psi_{\mathbf{x}_c}(t,\mathbf{x}) &= \frac{1}{\lambda}\int W(\vec{\mathbf{x}})\Sigma_c^{-1}(\mathbf{x})\mu(\mathbf{x}_1)P(\vec{\mathbf{x}})d\vec{\mathbf{x}}, \nonumber\\
&= \frac{1}{\lambda}E_P\left[W(\vec{\mathbf{x}})\Sigma_c^{-1}(\mathbf{x})\mu(\mathbf{x}_1)\right].
\end{align}
The optimal control (\ref{eq:opt_cont2}) is expressed as
\begin{align}
&\mathbf{u}^*(t,\mathbf{x})dt \nonumber\\
&= \frac{1}{\psi(t,\mathbf{x})}R^{-1}(\mathbf{x})G_c^T(\mathbf{x})\Sigma_c^{-1}(\mathbf{x})E_P\left[W(\vec{\mathbf{x}})\mu(\mathbf{x})dt\right],\nonumber\\
&=\frac{1}{\psi(t,\mathbf{x})}G_c^{-1}(\mathbf{x})B_c(\mathbf{x})E_P\left[W(\vec{\mathbf{x}})d\mathbf{w}^{(0)}\right], \label{eq:est_cont}
\end{align}
using $\mu(\mathbf{x})dt = B_c(\mathbf{x})d\mathbf{w}^{(0)}$ and $R^{-1}(\mathbf{x})G_c^T(\mathbf{x})\Sigma_c(\mathbf{x})^{-1} = G_c^{-1}(\mathbf{x})$.

The desirability function and the optimal control can be estimated from Monte-Carlo (MC) sampling procedure; the estimations for state $\mathbf{x}$ with $N$ sample trajectories are given by
\begin{equation}
\hat{\psi}(t,\mathbf{x}) = \frac{1}{N}\sum_{k=1}^Nw^k, \label{eq:est_cost2}
\end{equation}
and
\begin{equation}
\hat{\mathbf{u}}(t,\mathbf{x})\delta t = \frac{1}{N\hat{\psi}(t,\mathbf{x})}G_c^{-1}(\mathbf{x})B_c(\mathbf{x})\sum_{k=1}^N w^k \delta\mathbf{w}^k,
\end{equation}
where the weights, $w$, and the first Brownian increments, $\delta\mathbf{w}$, of the $k^\text{th}$ sample trajectory are obtained from following stochastic simulation.
Let $\delta t = (t_f-t)/I_s$ be sufficiently small time step for simulation of a continuous stochastic process and $I_s$ be simulation time steps.
\begin{enumerate}
	\item Set $i = 0,~\mathbf{X}_i = \mathbf{x}$.
	\item $\mathbf{X}_{i+1} = \mathbf{X}_i + \mathbf{f}(\mathbf{X}_i)\delta t + B(\mathbf{X}_i)\mathbf{Z}_i\sqrt{\delta t}$, where $\mathbf{Z}_i\sim N(0,I_m)$.
	\item If $i<I_s-1$, then $i=i+1$ and go to step 2.
	\item If $i=I_s-1$, then finish the simulation. \\Return $w^k = \exp\left(-\frac{1}{\lambda}(\phi(\mathbf{X}_{i+1})+\delta t\sum_{j=0}^iq(\mathbf{X}_j))\right)$ and $\delta\mathbf{w}^k=\mathbf{Z}_0\sqrt{\delta t}$.
\end{enumerate}

\subsection{Change of Measure (Importance Sampling)} \label{subsec:ch_mea}
In the naive MC sampling process, the sample trajectories for the estimation are collected from the passive diffusion dynamics (\ref{eq:dyn_passive}).
Most trajectories, however, may be useless (i.e. they hit the obstacle or reach the goal region through very awkward way, which are far from optimum), because they are driven only by white noise.
Rather than using naive MC sampling, it is possible to utilize advanced sampling technique to improve the quality of samples;
the importance sampling scheme is widely adopted in the path integral control literature.
Let $\mathbf{u}_{\text{in}}(t,\mathbf{x}) = g(t,\mathbf{x}(t))$ be any stationary or non-stationary policy, e.g., open loop control tape, trajectory tracking controller, etc.
Then, we can consider the new stochastic dynamics which drifts by the predefined (feedback) policy $\mathbf{u}_{\text{in}}(t,\mathbf{x}) = g(t,\mathbf{x}(t))$,
\begin{equation}
d\mathbf{x} = f(\mathbf{x})dt + G(\mathbf{x})\mathbf{u}_{\text{in}}dt + B(\mathbf{x})d\mathbf{w}^{(1)}, \label{eq:dyn_control2}
\end{equation}
and let $Q$ be a probability measure of the corresponding trajectories.

Then, the trajectories from the above stochastic dynamics can be used to estimate the desirability function and the optimal control, which is referred as a measure change or importance sampling.
Rewriting (\ref{eq:est_cost}) and (\ref{eq:est_cont}) yields
\begin{equation}
\psi(t,\mathbf{x}) = \int W(\vec{\mathbf{x}}) \frac{dP(\vec{\mathbf{x}})}{dQ(\vec{\mathbf{x}})}dQ(\vec{\mathbf{x}}) = E_Q\left[W(\vec{\mathbf{x}})\frac{dP(\vec{\mathbf{x}})}{dQ(\vec{\mathbf{x}})}\right], \label{eq:impor_sam1}
\end{equation}
and
\begin{equation}
\mathbf{u}^*(t,\mathbf{x})dt = \frac{1}{\psi(t,\mathbf{x})}G_c^{-1}(\mathbf{x})E_Q\left[W(\vec{\mathbf{x}})\mu(\mathbf{x})dt\frac{dP(\vec{\mathbf{x}})}{dQ(\vec{\mathbf{x}})}\right]. \label{eq:impor_sam2}
\end{equation}

The Radon-Nikodym derivative of $P$ with respect to $Q$, $\frac{dP(\vec{\mathbf{x}})}{dQ(\vec{\mathbf{x}})}$, can be obtained from following corollary.

\begin{cor}[Girsanov's Theorem \cite{gardiner1985handbook,theodorou2015nonlinear}]
	Suppose $P$ and $Q$ are the probability measures induced by the trajectories (\ref{eq:dyn_passive}) and (\ref{eq:dyn_control2}), respectively.
	Then the Radon-Nikodym derivative of $P$ with respect to $Q$, $\frac{dP(\vec{\mathbf{x}})}{dQ(\vec{\mathbf{x}})}$, is given by
	\begin{align}
	\frac{dP}{dQ} &= \exp(-\frac{1}{2\lambda}\int^{t_f}_t\mathbf{u}_{\text{in}}^T(\tau)G_c^T\Sigma_c^{-1}G_c\mathbf{u}_{\text{in}}(\tau)d\tau \nonumber\\
	&~~~~~~~~~~~~~~-\frac{1}{\lambda} \int^{t_f}_t\mathbf{u}_{\text{in}}^T(\tau)G_c^T\Sigma_c^{-1}B_cd\mathbf{w}^{(1)})\nonumber\\
	&= \exp(-\frac{1}{2\lambda}\int^{t_f}_t\mathbf{u}_{\text{in}}^T(\tau)R\mathbf{u}_{\text{in}}(\tau)d\tau \nonumber\\
	&~~~~~~~~~~~~~-\frac{1}{\lambda} \int^{t_f}_t\mathbf{u}_{\text{in}}^T(\tau)G_c^T\Sigma_c^{-1}B_cd\mathbf{w}^{(1)}),
	\end{align}
	with $\mathbf{u}_{\text{in}}(\tau) = g(\tau,\mathbf{x}(\tau))$ by a slight abuse of notation.
\end{cor}
With new probability measure $Q$, sampling procedure is changed as
\begin{enumerate}
	\item Set $i = 0,~t_i=t,~\mathbf{X}_i = \mathbf{x}$.
	\item $t_{i+1}=t_i+\delta t,~\mathbf{X}_{i+1} = \mathbf{X}_i + \mathbf{f}(\mathbf{X}_i)\delta t + G(\mathbf{X}_i)\mathbf{u}_{\text{in}}(t_i)\delta t + B(\mathbf{X}_i)\mathbf{Z}_i\sqrt{\delta t}$, where $\mathbf{u}_{\text{in}}(t_i) = g(t_i,\mathbf{X}_i)$ and $\mathbf{Z}_i\sim N(0,I_m)$.
	\item If $i<I_s-1$, then $i=i+1$ and go to step 2.
	\item If $i=I_s-1$, then finish the simulation. Return $w^k = \exp\left(-\frac{1}{\lambda}(\phi(\mathbf{X}_{i+1})+\delta t\sum_{j=0}^iL_j)\right)$ and $\delta\mu^k=\mathbf{u}_{\text{in}}(t_0)\delta t + \mathbf{Z}_0\sqrt{\delta t}$, where $L_j \equiv q(\mathbf{X}_j)+\frac{1}{2}\mathbf{u}_{\text{in}}^T(t_j)R\mathbf{u}_{\text{in}}(t_j)+\mathbf{u}_{\text{in}}^T(t_j)G_c^T\Sigma_c^{-1}B_c\mathbf{Z}_j/\sqrt{\delta t}$.
\end{enumerate}
The estimation of the desirability function is the same as (\ref{eq:est_cost2}) but because $\mu(\mathbf{x})dt = G_c(\mathbf{x})\mathbf{u}_{\text{in}}dt + B_c(\mathbf{x})d\mathbf{w}^{(1)}$ by substituting it to (\ref{eq:impor_sam2}), the estimation of the optimal control is given as,
\begin{align}
&\hat{\mathbf{u}}(t,\mathbf{x})\delta t  \nonumber\\
&=  \frac{1}{N\hat{\psi}(t,\mathbf{x})}\sum_{k=1}^N w^k\left(g(t,\mathbf{x})\delta t+G_c^{-1}(\mathbf{x})B_c(\mathbf{x})\delta\mathbf{w}^k\right) \nonumber\\
&= g(t,\mathbf{x})\delta t + \frac{1}{N\hat{\psi}(t,\mathbf{x})}G_c^{-1}(\mathbf{x})B_c(\mathbf{x})\sum_{k=1}^N w^k \delta\mathbf{w}^k. \label{eq:impor_con}
\end{align}
Note that all the estimations are unbiased~\cite{kappen2015adaptive,thijssen2015path}.
Especially, it is proven that the variance of estimation decreases as  $\mathbf{u}_{\text{in}}$ becomes closer to the real optimal control $\mathbf{u}^*$ \cite{thijssen2015path}.

\section{Topology-Guided Path Integral Control Algorithm}
\subsection{High-level Description of Proposed Algorithm}
By using importance sampling, sample trajectories are obtained around (or biased to) the \textit{reference trajectory} induced by the feedback policy with $\mathbf{u}_{\text{in}}=g(\mathbf{x})$.
Then through path integral procedure, the optimal trajectory/control is obtained by \textit{modifying} the reference trajectory/control.
However, the modification may be inaccurate if the amount of samples are not enough or may force the result to local optimum if the samples are far from global optimum.
Note that the problems addressed in this work may have many local optima, because the state space of the problem is highly non-convex because of obstacle regions.
The difficulty caused from non-convex space can be resolved if we have sample trajectories around every local optimum.

In this section, we propose the Path Integral with Fast Marching Homology-embedded Tree (PI-FMHT*) algorithm in order to resolve such difficulty.
The algorithm consists of expansion (Algorithm \ref{alg:FMHT}) and execution (Algorithm \ref{alg:Execution}) phases,
where the former operates in lead-time, and latter runs on-line; such construction has been utilized widely, e.g., in \cite{arslan2014information,jeon2015optimal}.
In expansion phase, the algorithm finds many different topological classes of trajectories for deterministic optimal motion planning problem.
And in execution phase, guided by feedforward and/or feedback policy induced by the motion plans, the optimal control input is computed in a receding horizon scheme with the path integral formula.

\subsection{Expansion phase: Sampling-based Algorithm for Topological Motion planning}
\subsubsection{Topological Representation of Trajectories in 2D}
Presence of obstacles in an environment differentiates topological classes among trajectories.
Suppose the configuration space, $\mathcal{C}$, and the obstacles are given by 2-dimensional subsets of $\mathbb{R}^2$.
Let $\sigma:[0,1]\rightarrow\mathbb{C}$ be a trajectory in the configuration space and $\sigma_1$ and $\sigma_2$ connecting the same start and end coordinates.
The two trajectories are called homologous if $\sigma_1$ together with $\sigma_2$ (the later with opposite orientation) forms the complete boundary of a 2-dimensional manifold embedded in $\mathcal{C}$ not containing/intersecting any of the obstacles \cite{bhattacharya2012topological}.

The configuration space can be represented as a subset of the complex plane $\mathbb{C}$, i.e. $(x,y)\in\mathcal{C} \Leftrightarrow x+iy\in\mathbb{C}$.
The obstacles are also represented as subsets of the complex plane, $\mathcal{O}_1, \mathcal{O}_2, ..., \mathcal{O}_N\subset\mathbb{C}$, and each obstacle has one \textit{representative point} which is denoted as $\zeta_l\in\mathcal{O}_l,~\forall l = 1,...,N$.
For a given set of representative points, the obstacle marker function, $\mathcal{F}:\mathbb{C}\rightarrow\mathbb{C}^N$, is defined as follows,
\begin{equation}
\mathcal{F}(z)=\left[\frac{1}{z-\zeta_1}, \frac{1}{z-\zeta_2}, \cdots, \frac{1}{z-\zeta_N}\right]^T.
\end{equation}
Then, we can define \textit{$H$-signature}, $\mathcal{H}_2:C_1(\mathbb{C})\rightarrow\mathbb{C}^N$, which represent homology class of trajectory as:
\begin{equation}
\mathcal{H}_2(\sigma)=\frac{1}{2\pi}\textit{Im}\left(\int_\sigma\mathcal{F}(z)dz\right),
\end{equation}
where $C_1(\mathbb{C})$ is the set of all curves/trajectories in $\mathbb{C}$.

Especially, when the trajectory from $z_1$ to $z_2$ is \textit{short} enough (that is, a straight line connecting the same points is in same homology class), its $H$-signature can be calculated analytically as
\begin{equation}
\left(\mathcal{H}_2(e)\right)_l = \frac{1}{2\pi}\absmin_{k\in\mathbb{Z}}(\arg(z_2-\zeta_l)-\arg(z_1-\zeta_l)+2k\pi),
\end{equation}
where function $\absmin$ returns the value which have the minimum absolute value.

If two trajectories $\sigma_1$ and $\sigma_2$ connecting the same points have the same $H$-signatures, $\mathcal{H}_2(\sigma_1)=\mathcal{H}_2(\sigma_2)$, they are homologous and the reverse is also true.
Also, we can restrict the homology class of trajectories by defining disjoint sets of allowed and blocked $H$-signature, $\mathcal{A}$ and $\mathcal{B}$, where $\mathcal{U}=\mathcal{A}\cup\mathcal{B}$ and $\mathcal{U}$ denotes the set of the $H$-signatures of all trajectories.
By well restricting the allowed $H$-signature set, the topological motion planning algorithm can secure scalability with the number of obstacles.
It can be observed from Fig. \ref{fig:c00} that there can be different trajectories which connect the same points and have different $H$-signatures.

The $H$-signatures for a higher dimensional space can be constructed by defining it directly \cite{bhattacharya2013invariants} or by using the configuration space mappings to 2-dimensional spaces \cite{pokorny2016high}.

\subsubsection{Sampling-based Algorithm for Topological Motion planning}
\begin{algorithm}[t]
	\caption{Expansion: FMHT* algorithm}\label{alg:FMHT}
	\begin{algorithmic}[1]
		\State $v_{goals} = \textsc{SampleGoal}(k_1);$
		\State $V\gets \{v_{goals}\}\cup\textsc{SampleFree}(k);~E\gets\emptyset;$
		\State $N_{open}\gets \{v_{goals}.n\};$
		\While{$\sim\textsc{Isempty}(N_{open})$}
		\State $z\gets\argmin_{n\in N_{open}}\{c(n)\};$
		\State $(V^b_z,E_{V^b_z,x(z)})\gets\textsc{NearBackward}(V,x(z))$
		\State $N_{temp} \gets \textsc{Propagate}(z, E_{V^b_z,x(z)});$
		\State $N_{near} \gets N_{temp}\setminus V^b_z.N;$
		\State $N_{open,new} \gets \emptyset;$
		\For {$n\in N_{near}$}
		\State $(V^f_n, E_{x(n),V^f_n})$
		\Statex $~~~~~~~~~\gets\textsc{Nearforward}(V,x(n));$
		\State $Y_{temp} \gets \textsc{Propagate}(n, E_{x(n),V^f_n});$
		\State $Y_{near} \gets Y_{temp}\cap V^f_n.N_{open};$
		\State $y_{min}\gets\argmin\limits_{y\in Y_{near}}\{c(y)+Cost(e_{x(n),x(y)})\}$
		\If{$\textsc{ObstacleFree}(e_{x(n),x(y_{min})})$}
		\State $n.c \gets y_{min}.c+Cost(e_{x(n),x(y_{min})});$
		\State $n.parent \gets y_{min};$
		\State $N_{open,new} \gets N_{open,new}\cup n;$
		\State $E \gets E\cup e_{x(n),x(y_{min})};$
		\EndIf
		\EndFor
		\State $N_{open} \gets (N_{open}\cup N_{open,new})\setminus \{z\};$
		\State $V\gets \textsc{AppendNode}(V, N_{open,new});$
		\EndWhile
		\State \Return{$T = (V, E)$}
	\end{algorithmic}
\end{algorithm}

This subsection is devoted to explain the expansion phase of PI-FMHT* algorithm, named FMHT*, which aims to find all optimal trajectories in different homology classes for deterministic approximation of the original problem.
FMHT* is batch-type algorithm like FMT*;
it generates $k$ samples in free configuration space.
Then from the nodes in the goal region, the $r$-disk graph is constructed and concurrently projects the tree into $H$-signature augmented space in order of cost-to-go.
With this outward moving, FMHT* performs the \textit{direct dynamic programming recursion} with \textit{lazy collision checking}.
The graph in state space is defined by a set of vertices, $V$, and edges, $E$, where each vertex is composed of a state, $v.x$, and set of associated nodes $v.N$.
Each node $n\in v.N$ has its $H$-signature, $n.H$, a cost, $n.c$, and a parent node $n.\text{parent}$.

FMHT* is shown in Algorithm \ref{alg:FMHT}.
Some required functions are described as follows:
\begin{itemize}
	\item \textsc{SampleGoal}($k$) function samples $k$ states from the goal region, $x_{goal}\in \mathcal{X}_{goal}$, and returns them by appending nodes into each vertex as $n(x) = x_{goal},~c(n) = 0,~par(n) = \emptyset$ and $H(n) = \mathcal{H}(e_{goal})$ where $\mathcal{H}(e_{goal})$ denotes $H$-signature of trajectory, $e_{goal}$, which is the straight line between $x_{goal}$ and the \textit{goal representative point}.
	\item \textsc{SampleFree}($k$) function returns $k$ random states from the free configuration space.
	\item \textsc{NearForward}$(V,x)$ and \textsc{NearBackward}$(V,x)$ functions return nearby vertices within a cost of $r_{|V|} = \gamma\left(\frac{\log|V|}{|V|}^{1/d}\right)$ (see \cite{janson2015fast}) among the set of vertices, $V$, from and to $x$, respectively, and also return corresponding optimal trajectories without considering obstacles; when the planning problem has \textit{kinodynamic} constraints, the optimal trajectory is the solution of two point boundary value problem, which can be computed in various ways according to the system dynamics and cost~\cite{webb2013kinodynamic,ha2013successive,karaman2010optimal}.
	\item \textsc{Propagate}$(n,v)$ returns the new node, $n_{new}$, which is created by propagating $n$ to the vertex $v$; the new node is given as $x(n_{new})=v$ and $n_{new}.H = n.H + \mathcal{H}(e)$, where $e$ denotes the piece-wise straight line from $v(n)$ to $v$; when $H$-signature of the new node is blocked (i.e., $H(n_{new})\in\mathcal{B}$), the function does not return the new node.
	\item \textsc{ObstacleFree}$(e)$ takes a trajectory $e$ as an argument and checks whether it lies in obstacle free region or not.
	\item \textsc{AppendNode}$(V,N)$ adds nodes, $N$, to each vertex in $V$.
\end{itemize}

The algorithm operates as follows.
It first creates the node and vertex in the goal region then samples a set of states on the free configuration space, $\mathcal{X}_{free}$ (line 1--2).
Then the goal nodes are added to the open set and one of them is chosen as the minimum cost open node (line 3 and 5).
In the main loop, the algorithm finds backward near vertices of $x(z)$ and propagates the backward near vertices to $z$ in order to make candidates of new nodes, $N_{near}$ (line 6--7).
Then, it checks the nodes already exist in the tree and excludes the existing nodes from $N_{near}$ (line 8).
For each candidate node, the algorithm finds \text{open} forward near nodes in the tree (line 11--13) and finds the optimal connections without considering obstacles (line 14).
This procedure represents \textit{direct dynamic programming recursion} on $r_{|V|}$-disk graph and guarantees the optimal connection of the tree in obstacle free space from the fact that every new node must pass through a open node.
Then, the new node and edge are added to the tree if the connection is collision-free (line 15--19); if such connection is not collision-free, adding the new node is postponed.
This \textit{lazy collision checking} may induces sub-optimality of the tree but the number of costly collision checking is dramatically reduced;
also, it is known that the cases where a suboptimal connection is made become vanishingly rare as the number of samples increases~\cite{janson2015fast}.
After trying to make all connections to $N_{near}$, $z$ is excluded from the open set and $N_{open,new}$ is added to the open set and to the tree (line 22--23).
Then, the minimum cost open node, $z$, is chosen among the open set (line 24).
The algorithm proceeds to the next iteration by the minimum cost open node (line 5) unless the set of open nodes, $N_{open}$, is empty, and it returns the tree when the iteration ends.

\begin{figure*}[t]
	\centering
	\subfigure[]{
		\includegraphics*[width=.4\columnwidth,viewport= 105 30 315 285]{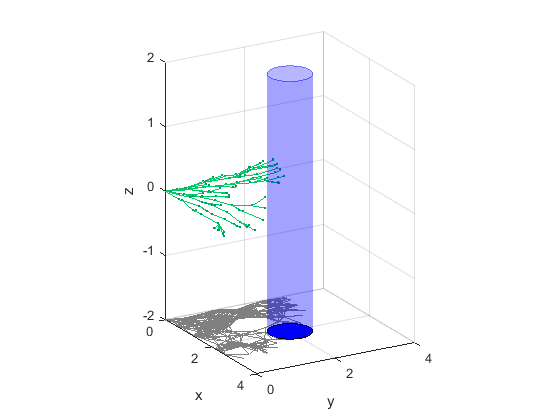}}
	\subfigure[]{
		\includegraphics*[width=.4\columnwidth,viewport=105 30 315 285]{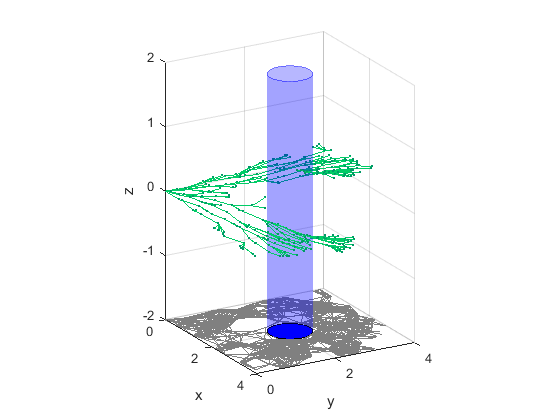}}
	\subfigure[]{
		\includegraphics*[width=.4\columnwidth,viewport=105 30 315 285]{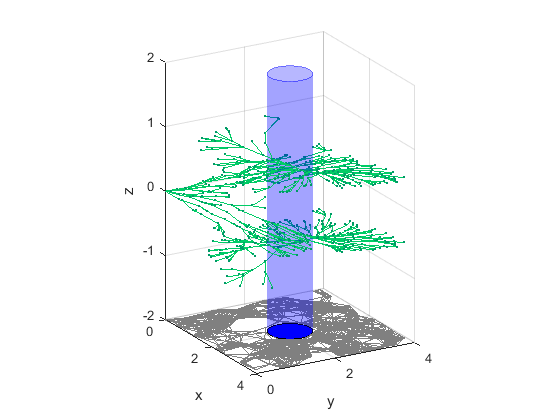}}
	\subfigure[]{
		\includegraphics*[width=.4\columnwidth,viewport=105 30 315 285]{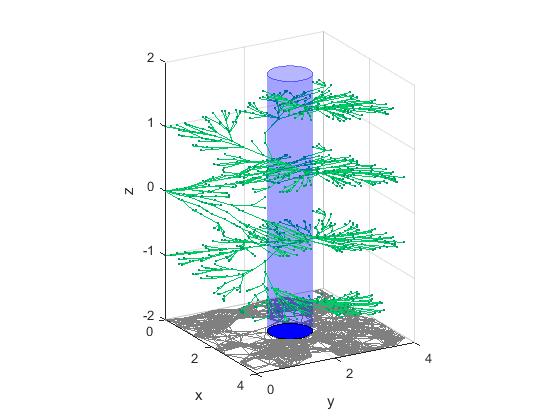}}
	\subfigure[]{
		\includegraphics*[width=.4\columnwidth,viewport= 105 30 315 285]{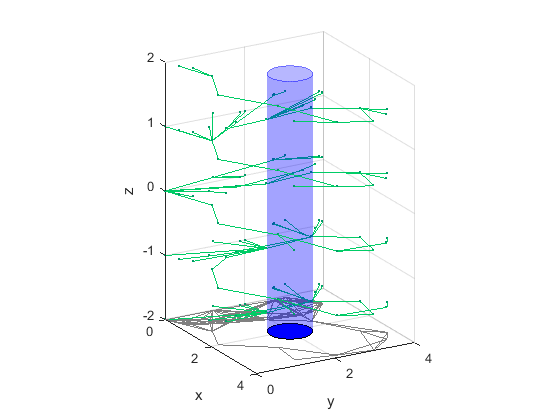}}
	\subfigure[]{
		\includegraphics*[width=.4\columnwidth,viewport=105 30 315 285]{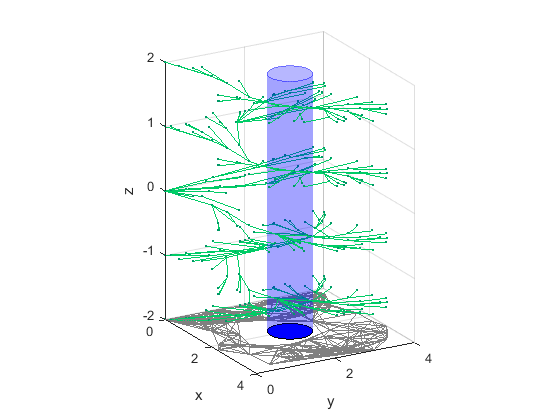}}
	\subfigure[]{
		\includegraphics*[width=.4\columnwidth,viewport=105 30 315 285]{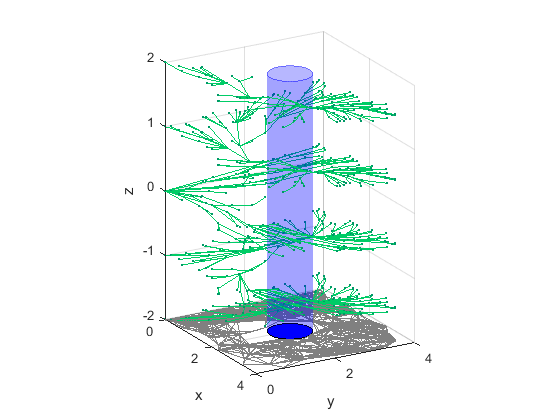}}
	\subfigure[]{
		\includegraphics*[width=.4\columnwidth,viewport=105 30 315 285]{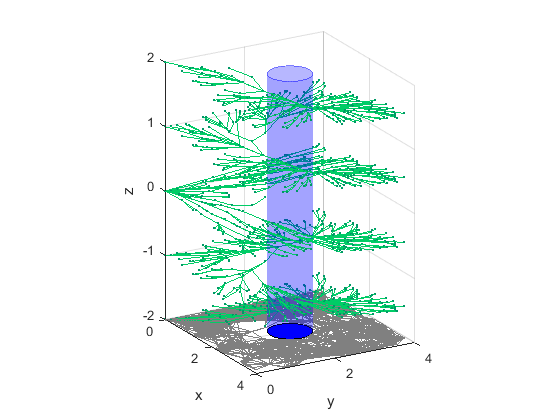}}
	\caption{The growth of the tree and the graph in the $H$-signature augmented space with (a) 100, (b) 300, (c) 600 and (d) 1000 nodes by the FMHT* algorithm and (e) 108, (f) 301, (g) 601 and (h) 1001 nodes by the RRHT* algorithm.}
	\label{fig:result1}
\end{figure*}
\begin{figure*}[t]
	\centering
	\subfigure[FMHT*]{
		\includegraphics*[width=.85\columnwidth]{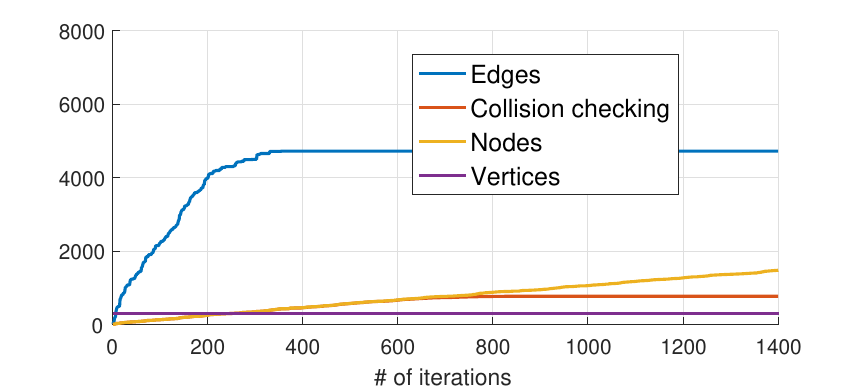}
		\label{fig:result2a}}
	\subfigure[RRHT*]{
		\includegraphics*[width=.85\columnwidth]{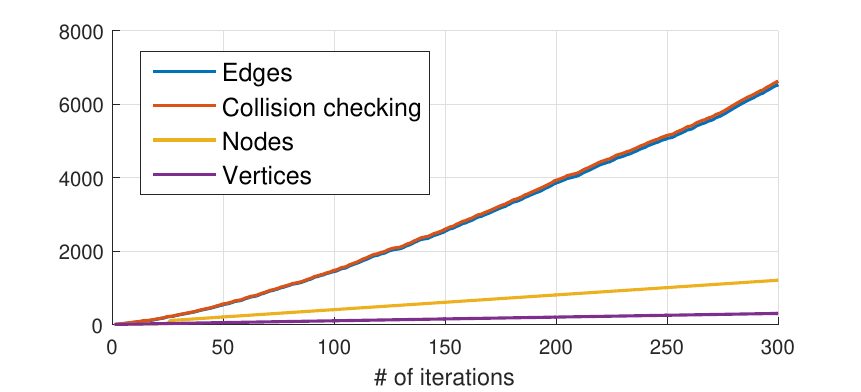}
		\label{fig:result2b}}
	\caption{The number of the edges in the graph, the collision checking, the nodes and the vertices w.r.t the number of iterations}
	\label{fig:result2}
\end{figure*}
\subsection{Execution Phase: Receding Horizon Path Integral Control}
\begin{algorithm}
	\caption{Execution: Receding Horizon Path Integral Control}\label{alg:Execution}
	\begin{algorithmic}[1]
		\State Given the current state $\mathbf{x}_{cur}$ and the Tree $(V,E);$
		\While{$\mathbf{x}_{cur}\notin \mathcal{X}_{goal}$}
		\State $\vec{\mathbf{X}}\gets\textsc{ExtractReference}(\mathbf{x}_{cur}, (V,E));$
		\For {$h \in \{1,...,H\}$}
		\State $g^{(h)}(t,\mathbf{x}) \gets \textsc{Controller}\left(\vec{\mathbf{x}}^{(h)}\right);$
		\State $\{\hat{\mathbf{u}}_i^{(h)}\delta t\}_{i=0,...,I_{RH}-1} $
		\Statex $~~~~~~~~~~\gets\textsc{PathIntegral}(\mathbf{x}_{cur},g^{(h)});$
		\EndFor
		\State $\{\hat{\mathbf{u}}_i\delta t\}_{i=0,...,I_{RH}-1} $
		\Statex$~~~~~\gets \{\frac{1}{H}\sum_{h=1}^H\hat{\mathbf{u}}^{(h)}\delta t\}_{i=0,...,I_{RH}-1};$
		\State $\mathbf{x}_{cur} \gets \textsc{ApplyControl}(\mathbf{x}_{cur}, \{\hat{\mathbf{u}}_i\delta t\}_{i=0,...,I_{RH}-1});$
		\EndWhile
	\end{algorithmic}
\end{algorithm}\vspace*{-4mm}
\begin{algorithm}
	\caption{ExtractReference($\mathbf{x}_{cur}, (V,E)$)}\label{alg:ExtractReference}
	\begin{algorithmic}[1]
		\State $(v_{new},E_{new}) \gets \textsc{ChooseParent}(V, \mathbf{x}_{cur});$
		\State $V \gets V \cup v_{new};~E \gets E \cup E_{new};$
		\State $\{\vec{\mathbf{x}}^{(h)},~h=1,2,...,H\}\gets$\textsc{ReconstructPath}($G\gets(V, E),v_{new}$);
		\Comment look at its ancestry to find the paths (node$\rightarrow$parent$\rightarrow$parent$\rightarrow$parent..., etc)
		\State \Return{$\vec{\mathbf{X}}\gets\{\vec{\mathbf{x}}^{(h)},~h=1,2,...,H\}$}
	\end{algorithmic}
\end{algorithm}
The execution phase of PI-FMHT* presented in Algorithm \ref{alg:Execution} computes and executes the optimal control for stochastic problem in a receding horizon fashion.
It consists of four procedures:
$\textsc{ExtractReference}(\mathbf{x}_{cur}, (V,E))$ shown in Algorithm \ref{alg:ExtractReference} takes the current state $\mathbf{x}_{cur}$ and the tree $(V, E)$ constructed from Algorithm \ref{alg:FMHT} as arguments and returns a set of all the allowed homology trajectories from $\mathbf{x}_{cur}$ to the roots of the tree in $\mathcal{X}_{goal}$.
Next, for each trajectory in the set, the controller, $g^{(h)}$, that makes a robot follow the trajectory is constructed (as is mentioned, the controller can be a simple open-loop control sequence or a tracking controller for the trajectory.).
Then in $\textsc{PathIntegral}(\mathbf{x}_{cur},g^{(h)})$, trajectories are sampled around each homology class and the optimal control is computed.
The time horizon considered in this procedure can be given by a user or set as the horizon of the minimum length trajectory among $\vec{\mathbf{X}}$.
Suppose there are $H$ number of stochastic dynamics (\ref{eq:dyn_control2}) controlled by $\mathbf{u}_{\text{in}}^{(h)}=g^{(h)}$ and let $Q_h,~h=1,2,...,H$ be corresponding probability measures.
Equations (\ref{eq:impor_sam1}) and (\ref{eq:impor_sam2}) can be rewritten as
\begin{equation}
\psi(t,\mathbf{x}) = \frac{1}{H}\sum_{h=1}^{H}E_{Q_h}\left[W(\vec{\mathbf{x}})\frac{dP(\vec{\mathbf{x}})}{dQ_h(\vec{\mathbf{x}})}\right],
\end{equation}
and
\begin{align}
&\mathbf{u}^*(t,\mathbf{x})dt \nonumber\\
&= \frac{1}{H}\sum_{h=1}^{H}\frac{1}{\psi(\mathbf{x})}G_c^{-1}(\mathbf{x})E_{Q_h}
\left[W(\vec{\mathbf{x}})\mu_h(\mathbf{x})dt\frac{dP(\vec{\mathbf{x}})}{dQ_h(\vec{\mathbf{x}})}\right].
\end{align}
This procedure can be viewed that, instead of using one trajectory distribution with $Q$, the \textit{mixture} of $H$ trajectory distributions is considered as a proposal distribution for the importance sampler.	
Suppose we sample $N$ trajectories from each homology class, $h = 1, 2, ..., H$, by procedure described in Section \ref{subsec:ch_mea} and let the weights of $k^\text{th}$ sample trajectory in $h^\text{th}$ homology class be indexed by $w^{(k,h)}$.
Then we have
\begin{equation}
\hat{\psi}(t,\mathbf{x}) = \frac{1}{H}\sum_{h=1}^H\hat{\psi}^{(h)}(t,\mathbf{x}), \label{eq:est_val_h}
\end{equation}
and
\begin{equation}
\hat{\mathbf{u}}(t,\mathbf{x})\delta t = \frac{1}{H}\sum_{h=1}^H\hat{\mathbf{u}}^{(h)}(t,\mathbf{x})\delta t, \label{eq:est_con_h}
\end{equation}
where $\hat{\psi}^{(h)}(t,\mathbf{x})\equiv \frac{1}{N}\sum_{k=1}^Nw^{(k,h)}$ and
\begin{align}
&\hat{\mathbf{u}}^{(h)}(t,\mathbf{x})\delta t \nonumber\\
&\equiv g^{(h)}(t,\mathbf{x})\delta t+\frac{1}{N\hat{\psi}(t,\mathbf{x})}\sum_{k=1}^N w^{(k,h)}G_c^{-1}(\mathbf{x})B_c(\mathbf{x})\delta\mathbf{w}^{(k,h)}. \label{eq:uh}
\end{align}
Note that from the above equations, the optimal control is only computed at the current time, $t$, and state, $\mathbf{x}_{cur}$.
However, if the control policy we want to compute is restricted as the open loop formulation, i.e., for $i=0,1,...,I_{RH}-1$
$$\mathbf{u}(\tau,\mathbf{x}) = \hat{\mathbf{u}}_i,~\forall \tau\in\left[t+i\times\delta t, t+(i+1)\times\delta t\right),$$ the state dependence term can be dropped and we can obtain the \textit{open loop control sequence} by storing $\delta\mu_i^k=g^{(h)}(t_i^k,X_i^k)\delta t+G_c^{-1}(X_i^k)B_c(X_i^k)\delta\mathbf{w}_i^{(k,h)}~\forall i=0,1,2,...,I_{RH}-1$ in the importance sampling procedure and using
\begin{align}
&\hat{\mathbf{u}}_i\delta t = \frac{1}{H}\sum_{h=1}^H\hat{\mathbf{u}}^{(h)}_i\delta t,\\
&\hat{\mathbf{u}}^{(h)}_i\delta t \equiv\frac{1}{N\hat{\psi}(\mathbf{x})}\sum_{k=1}^N w^{(k,h)}\delta\mu_i^k,~\forall i=0,1,...,I_{RH}-1,
\end{align}
rather than only storing $\delta\mathbf{w}^k=\mathbf{Z}_0\delta t$ and using \eqref{eq:uh}~(see \cite{thijssen2015path,theodorou2015nonlinear}).
As a result, $\textsc{PathIntegral}$ procedure computes the open loop control policy for one-period of receding horizon, $\tau\in\left[t,t+I_{RH}\delta t\right)$.
Such control is applied to the system for one-period by $\textsc{ApplyControl}$, then the overall algorithm repeats again until the state reaches the boundary of the domain.

\section{Comparison with Other Topological Motion Planners}
There have been some recent works on developing sampling-based algorithms for optimal topological motion planning: Winding-Augmented RRT* (WA-RRT*) \cite{pokorny2016high} and Rapidly-exploring Random Homology-embedded Tree star (RRHT*)~\cite{ha2016topology}.
FMHT*, WA-RRT*, and RRHT* inherit the properties of the FMT* and RRT* algorithms, respectively;
like FMT*, FMHT* is batch processing algorithm and performs the direct dynamic programming process and lazy collision checking which dramatically accelerates the speed of the algorithm \cite{janson2015fast};
WA-RRT* and RRHT* are incremental anytime algorithm like RRT* which finds a feasible trajectory quickly by rapidly exploring the configuration space and refines the solution for allowed computation time.
All algorithms are tailored to disk-connected graphs, where for the given connection radius, two vertices are considered as neighbor, and concurrently perform graph construction and graph search;
latter is key feature of sampling-based algorithm improving the scalability to a high-dimensional configuration space, because it makes the algorithms not suffer from the curse of dimensionality (the algorithms need not discretize the configuration space in advance).
While WA-RRT* samples a value of $H$-signature after sampling a configuration coordinate, FMHT* and RRHT* do not have \textit{additional} sampling step; they expand a graph directly in the configuration space and project an associated tree onto the \textit{H-signature augmented space}.
As a result, the FMHT* and RRHT* algorithms share the edge information for every layer of $H$-signature space and thus have potential to significantly reduce the computational cost caused by edge computation and its collision checking which are the computational bottleneck in many cases.

Because WA-RRT* and RRHT* are almost same in the other aspects (i.e., except additional $H$-signature sampling), we only compare the properties of FMHT* and RRHT* algorithms here.
To do so, a simple 2-dimensional configuration space with one obstacle is considered.
Fig. \ref{fig:result1} shows how the trees are expanded into the $H$-augmented space by the proposed algorithms.
Green lines represent the edges of the tree and dark-gray lines on the bottom denote the edge of the graph which the tree is projected by;
$x$ and $y$ axis denote the configuration, $z$ represents $H$-signature and the goal augmented state is $[0,0,0]^T$;
in this example, $H$-signature is scalar because there is only one obstacle.
It is shown in the top row of Fig. \ref{fig:result1} that the tree of FMHT* is expanded in order of the cost-to-go.
Also, the graph is expanded only in the early phase of algorithm and the tree is projected only by the expanded graph;
this implies that the algorithm does not need to compute the edges and check whether they collide or not when the tree is expanded to other $H$-signature layers (see Fig. \ref{fig:result2a}).
On the other hand, The bottom row of Fig. \ref{fig:result1} shows that RRHT* rapidly expands the tree to the whole space and rewires it.
In addition, note that RRHT* also shares the edge information (shown as the graph) through all $H$-signature layers;
it is also shown that the graph (vertices) projects the tree (nodes) into the augmented space (see Fig. \ref{fig:result2b}).
Finally, because the topological motion planner is operated in lead-time (i.e., off-line phase), a batch processing algorithm, FMHT*, is much more suitable to the proposed topology-guided path integral control framework.

\begin{figure*}[t]
	\centering
	\subfigure[]{
		\includegraphics*[width=.5\columnwidth]{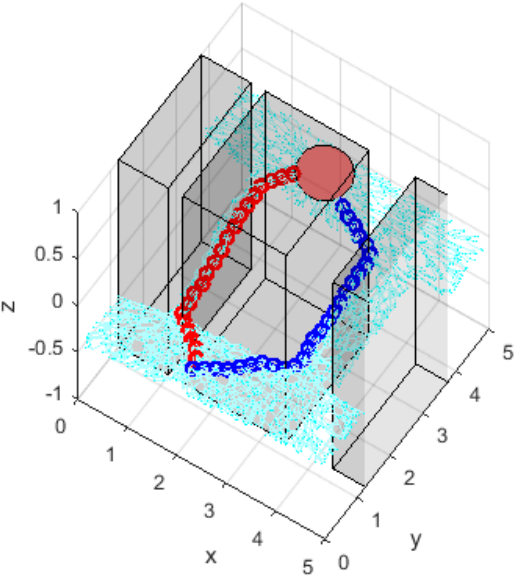}}
	\subfigure[]{
		\includegraphics*[width=.5\columnwidth]{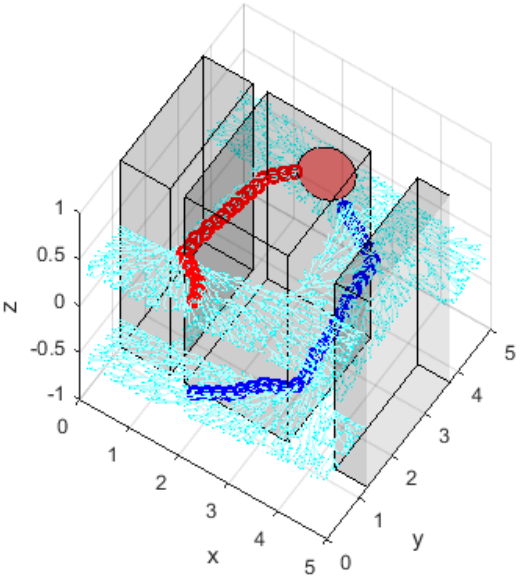}}
	\subfigure[]{
		\includegraphics*[width=.5\columnwidth]{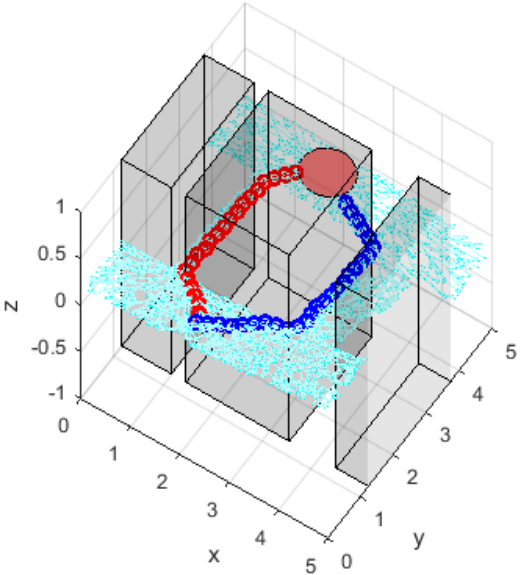}}
	\caption{The tree depicted by cyan edges is constructed by Expansion phase of PI-FMHT* (Algorithm \ref{alg:FMHT}) on $H$-signature augmented space for the single integrator example; a red circle represents goal region. $z$ axis denotes the values of $H$-signature with respect to the obstacle on (a) left side, (b) middle, and (c) right side. The circled solid lines colored by red and blue result from \textsc{ExtractReference}() and show the reference trajectories in different homology classes.}
	\label{fig:c00}
\end{figure*}
\section{Numerical Experiments}
\subsection{Drunk Spider: Choosing Slit}
\begin{figure*}[t]
	\centering
	\subfigure[$b=0.1$]{
		\includegraphics*[width=.45\columnwidth]{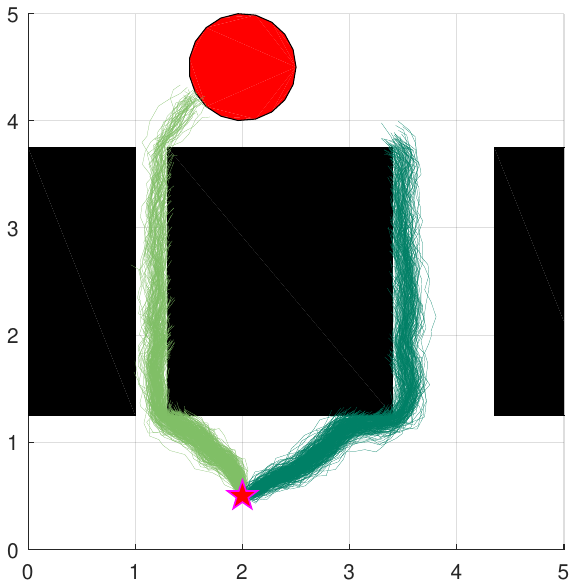}}
	\subfigure[$b=0.1$]{
		\includegraphics*[width=.45\columnwidth]{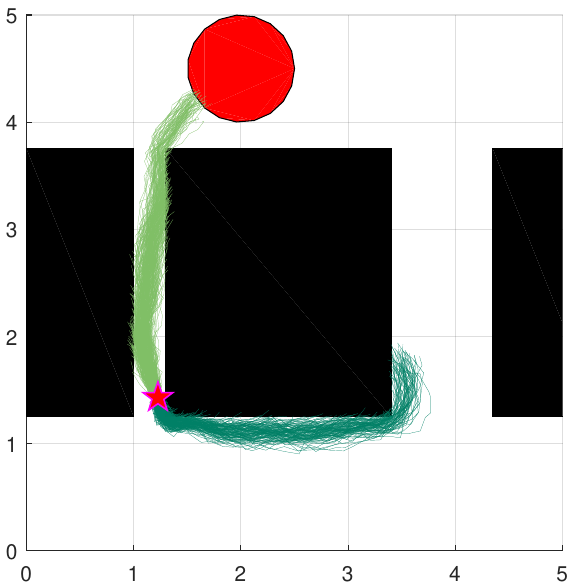}}
	\subfigure[$b=0.1$]{
		\includegraphics*[width=.45\columnwidth]{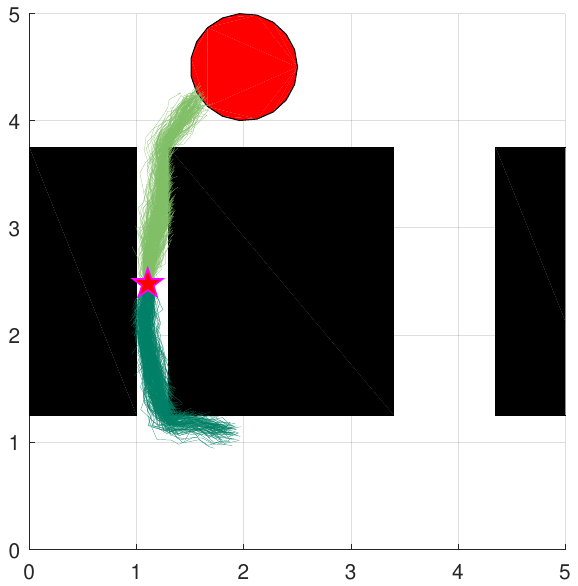}}
	\subfigure[$b=0.1$]{
		\includegraphics*[width=.45\columnwidth]{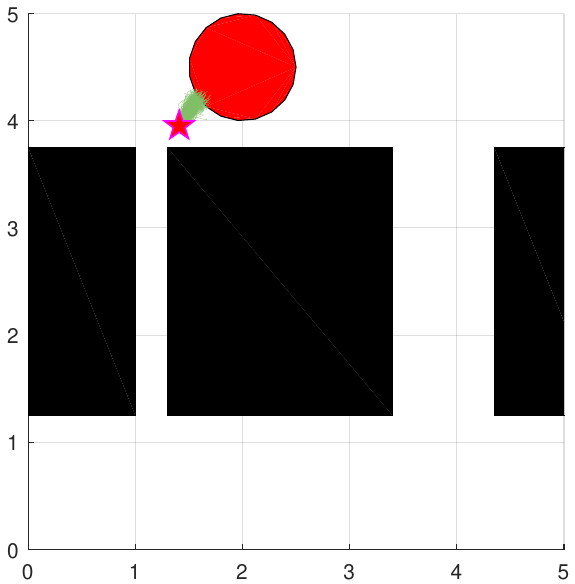}}
	\subfigure[$b=0.3$]{
		\includegraphics*[width=.45\columnwidth]{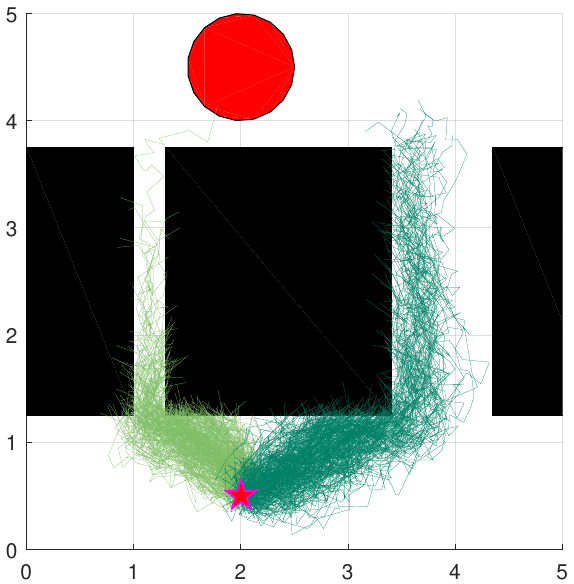}}
	\subfigure[$b=0.3$]{
		\includegraphics*[width=.45\columnwidth]{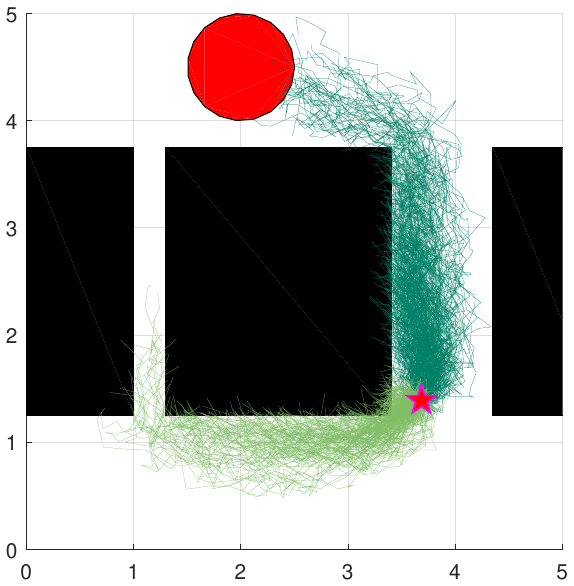}}
	\subfigure[$b=0.3$]{
		\includegraphics*[width=.45\columnwidth]{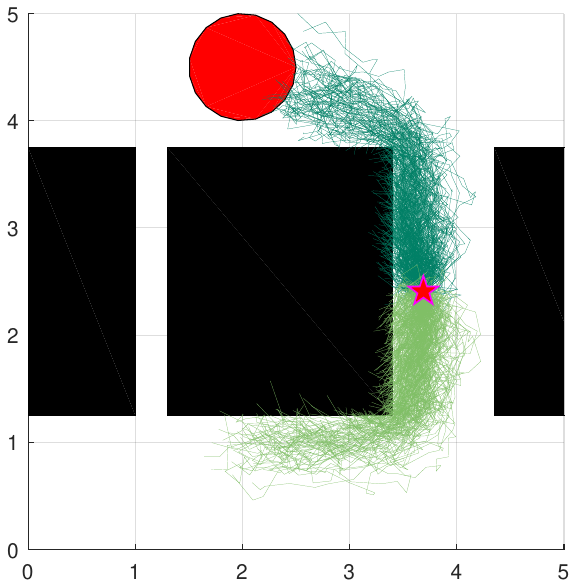}}
	\subfigure[$b=0.3$]{
		\includegraphics*[width=.45\columnwidth]{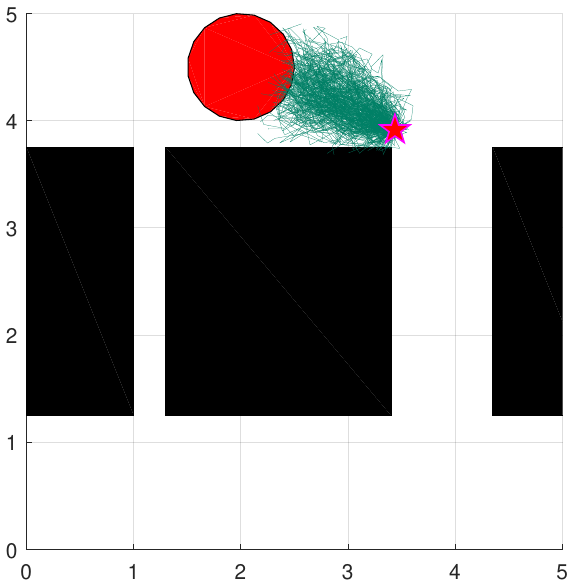}}
	\caption{Snapshots of execution phase of PI-FMHT* (Algorithm \ref{alg:Execution}) for the single integrator example, where $B=bI_2$ with (a)-(d) $b=0.1$ and (e)-(h) $b=0.3$. Colors of yellow, dark and bright green distinguish different homology classes, where thin edges and small-circled line represent the sample trajectories and the corresponding reference, respectively.}
	\label{fig:c01}
\end{figure*}	

\begin{figure*}[t]
	\centering	
	\subfigure[$b=0.05$]{
		\includegraphics*[width=.4\columnwidth,viewport= 85 30 350 300]{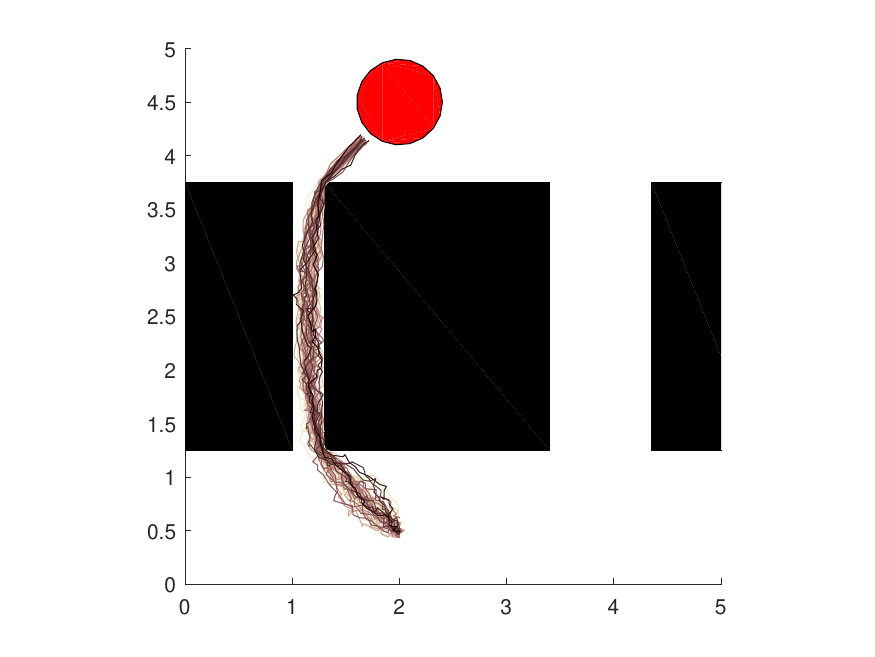}}
	\subfigure[$b=0.1$]{
		\includegraphics*[width=.4\columnwidth,viewport= 85 30 350 300]{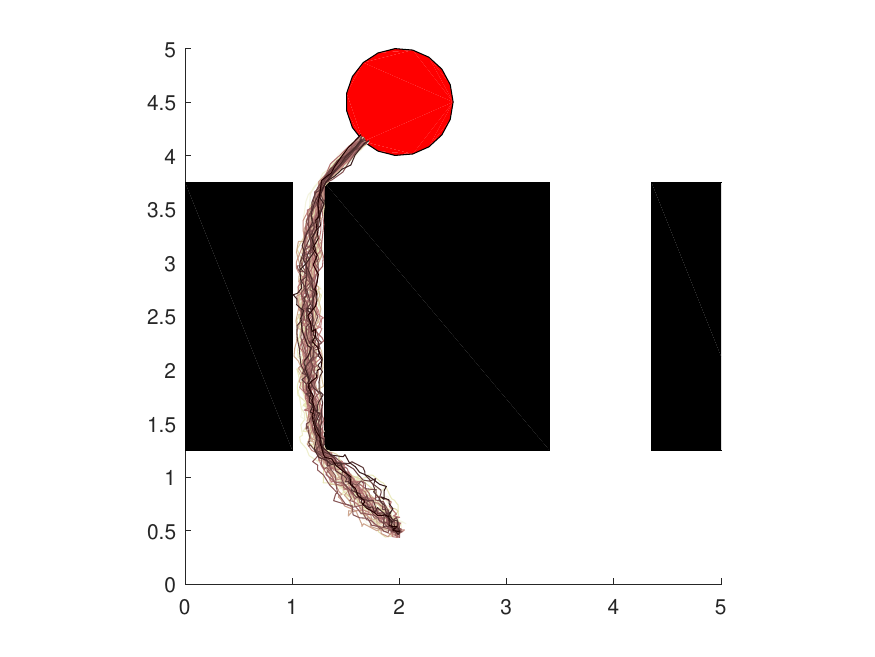}}
	\subfigure[$b=0.3$]{
		\includegraphics*[width=.4\columnwidth,viewport= 85 30 350 300]{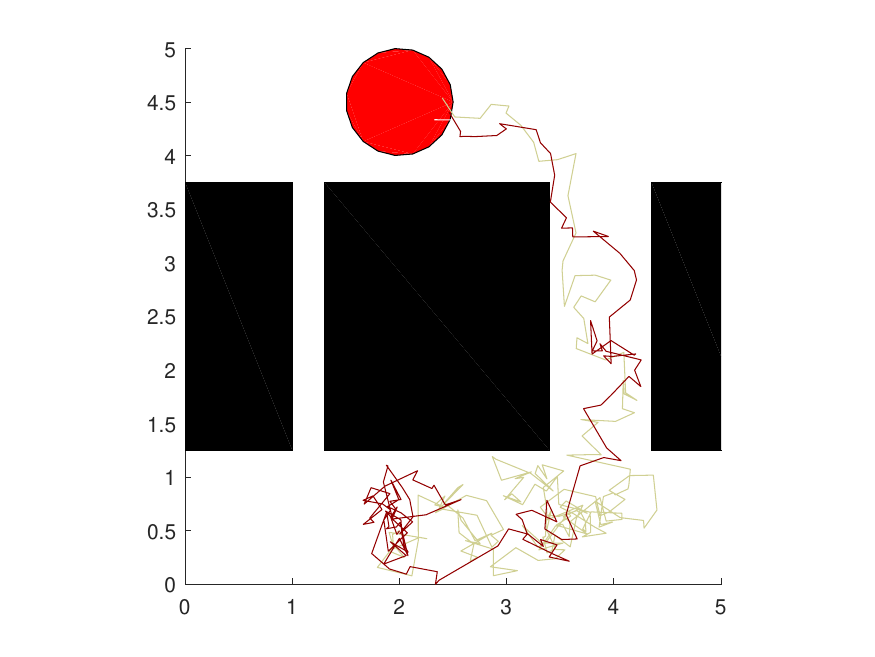}}
	\subfigure[$b=0.5$]{
		\includegraphics*[width=.4\columnwidth,viewport= 85 30 350 300]{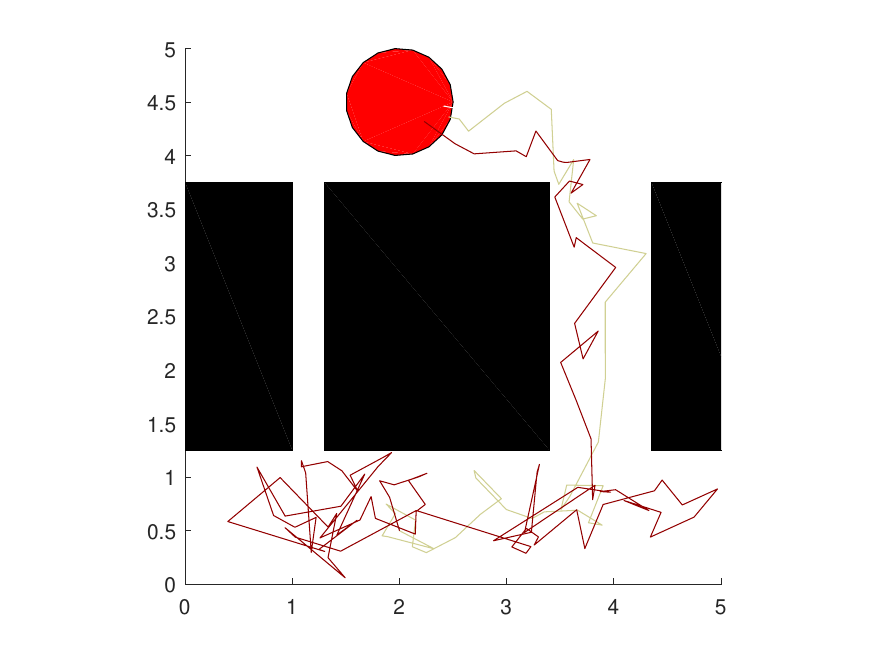}}
	
	\subfigure[$b=0.05$]{
		\includegraphics*[width=.4\columnwidth,viewport= 85 30 350 300]{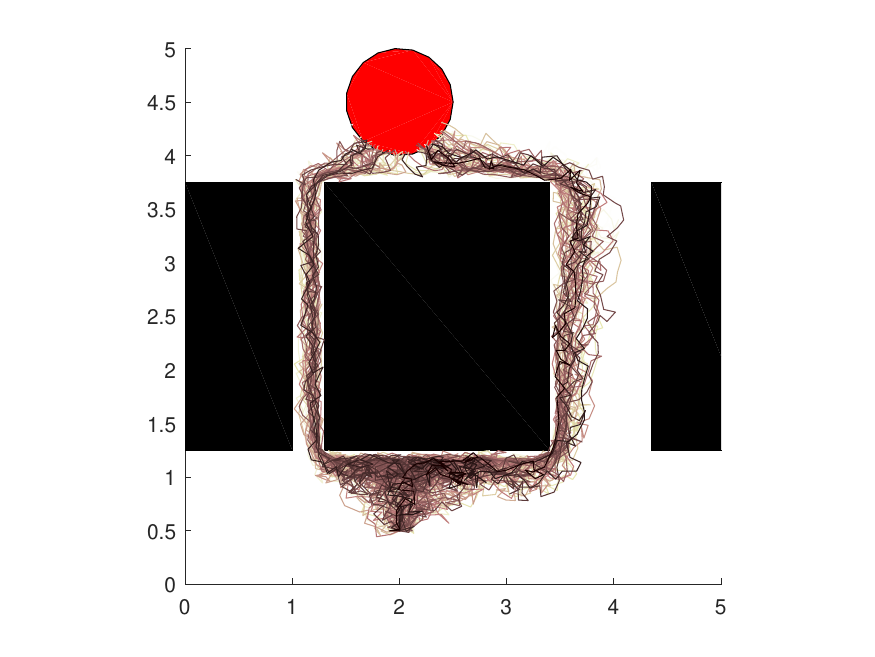}}
	\subfigure[$b=0.1$]{
		\includegraphics*[width=.4\columnwidth,viewport= 85 30 350 300]{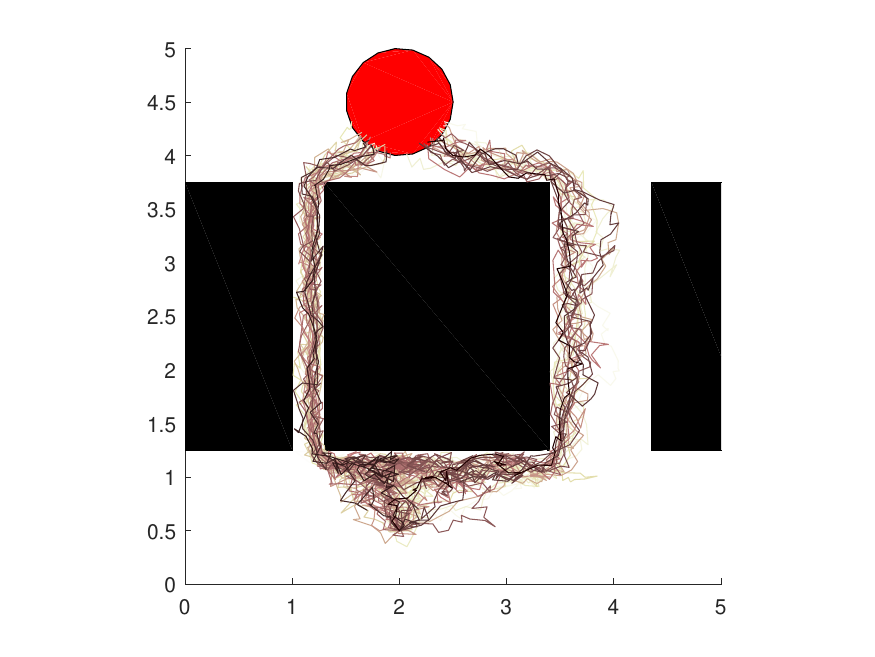}}
	\subfigure[$b=0.3$]{
		\includegraphics*[width=.4\columnwidth,viewport= 85 30 350 300]{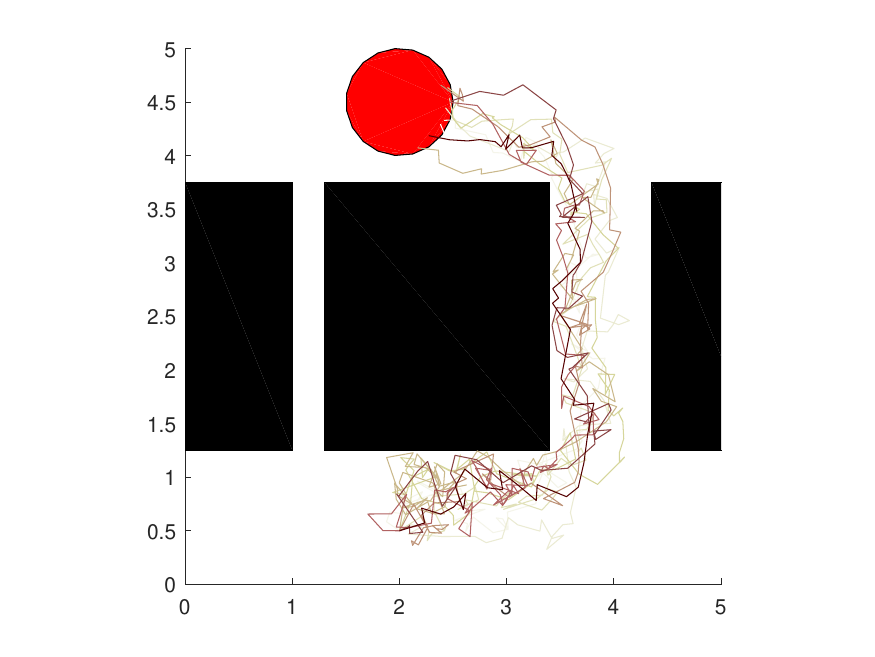}}
	\subfigure[$b=0.5$]{
		\includegraphics*[width=.4\columnwidth,viewport= 85 30 350 300]{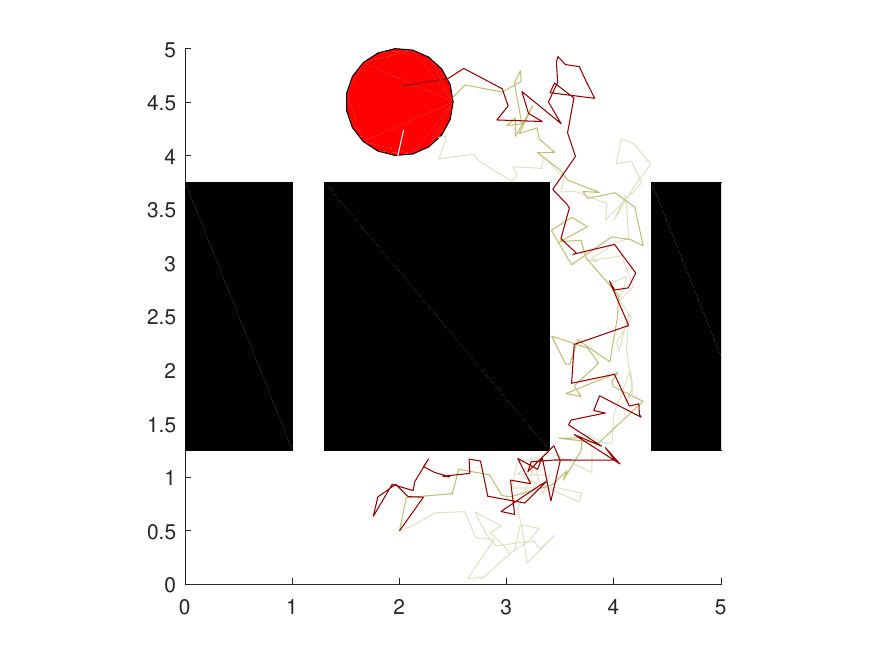}}
	
	\subfigure[$b=0.05$]{
		\includegraphics*[width=.4\columnwidth,viewport= 85 30 350 300]{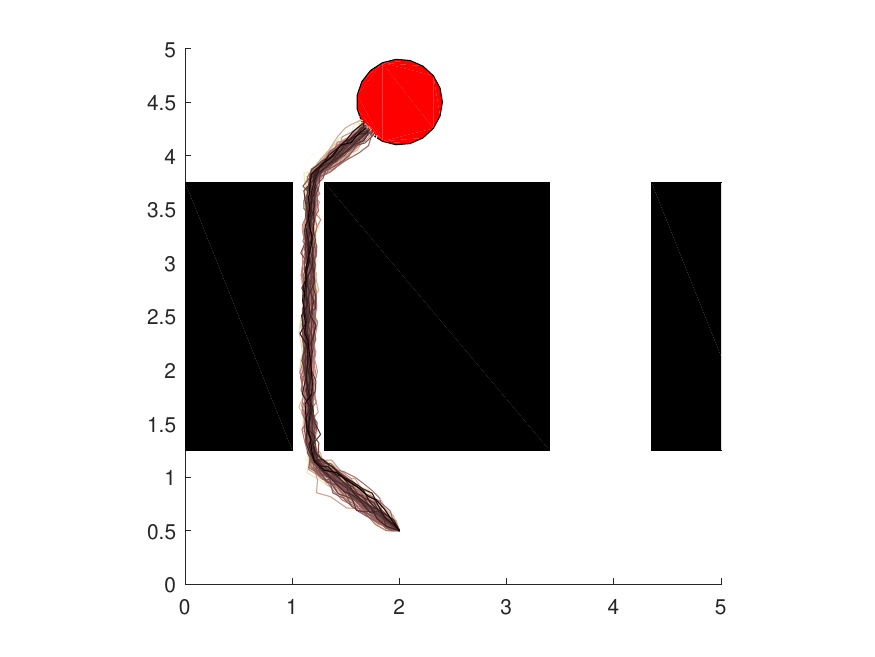}}
	\subfigure[$b=0.1$]{
		\includegraphics*[width=.4\columnwidth,viewport= 85 30 350 300]{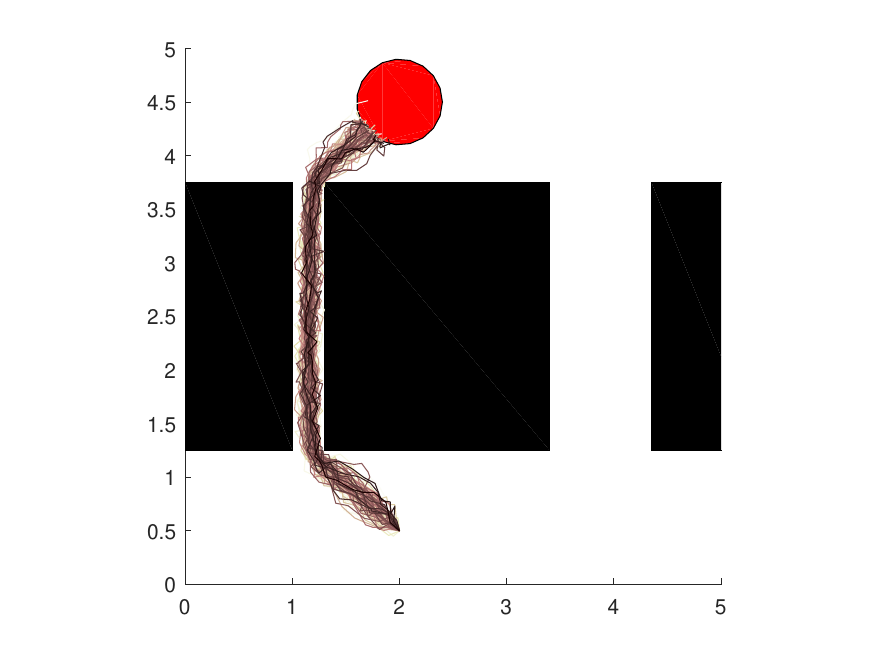}}
	\subfigure[$b=0.3$]{
		\includegraphics*[width=.4\columnwidth,viewport= 85 30 350 300]{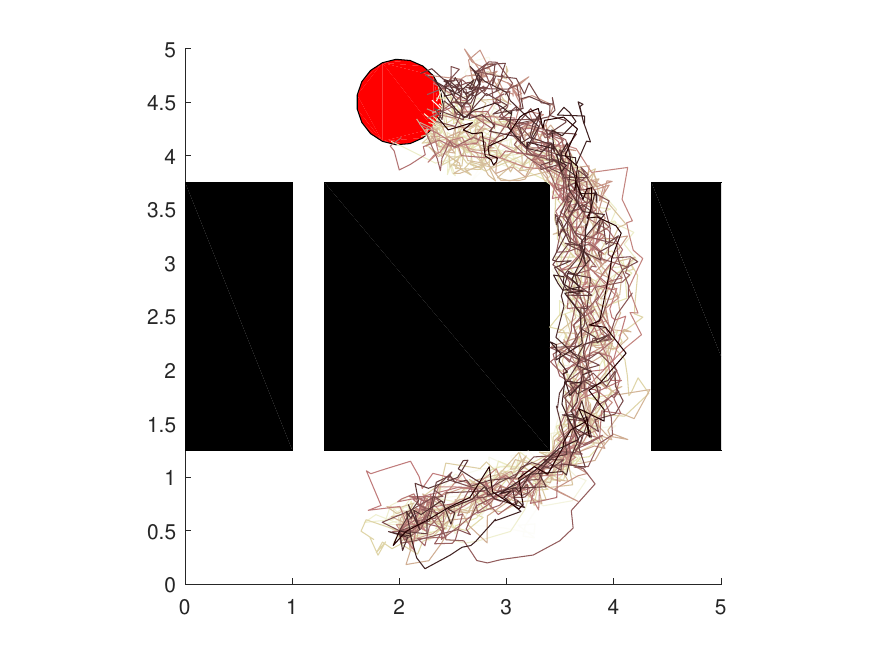}}
	\subfigure[$b=0.5$]{
		\includegraphics*[width=.4\columnwidth,viewport= 85 30 350 300]{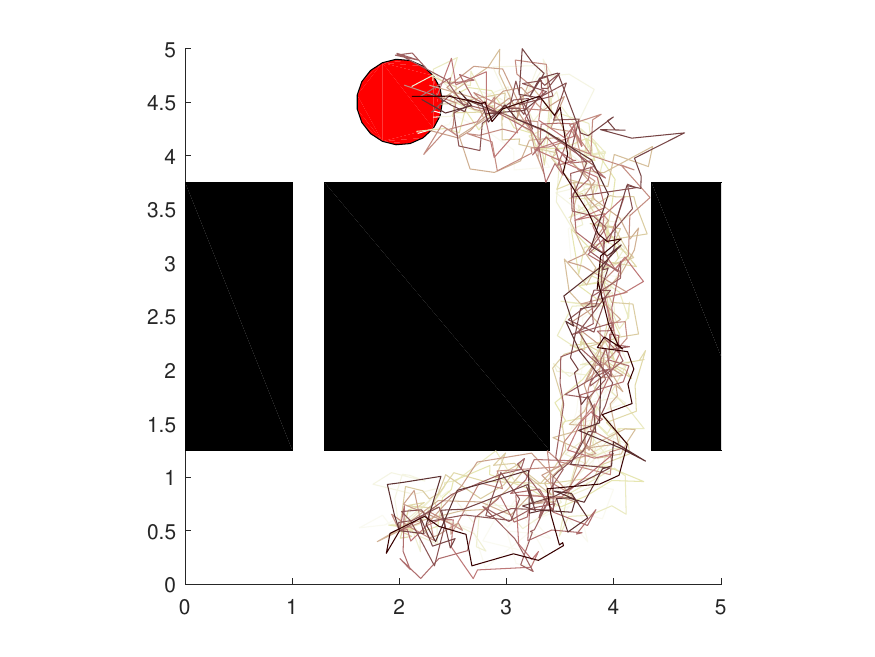}}
	\caption{Resulting collision free trajectories from (a-d) iterative PI, (e-h) PI-RRT, and (i-l) PI-FMHT*.}
	\label{fig:c03}
\end{figure*}

\begin{figure}[t]
	\centering	
	\subfigure[$K=1$]{
		\includegraphics*[width=.45\columnwidth,viewport= 85 30 350 300]{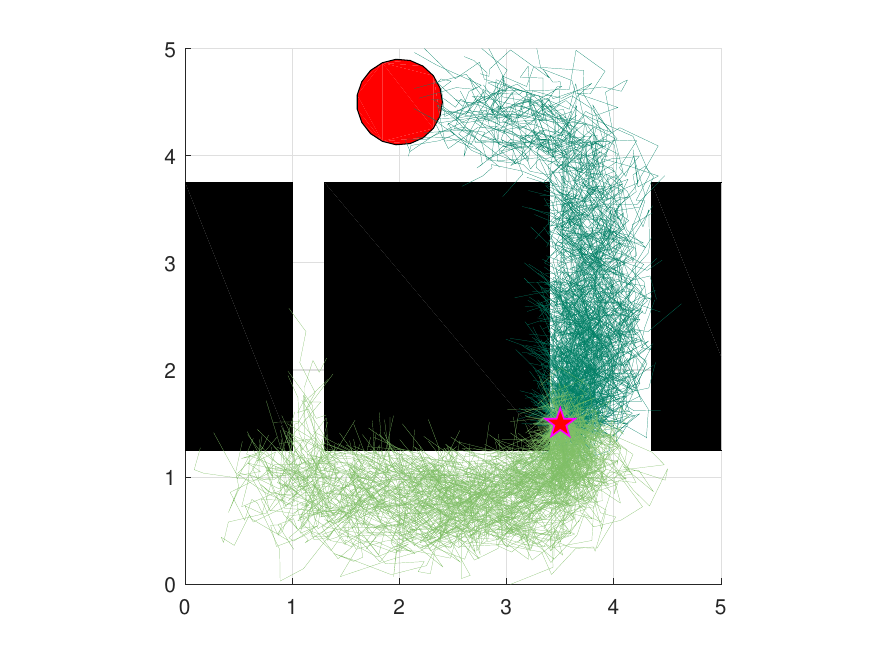}}
	\subfigure[$K=0$]{
		\includegraphics*[width=.45\columnwidth,viewport= 85 30 350 300]{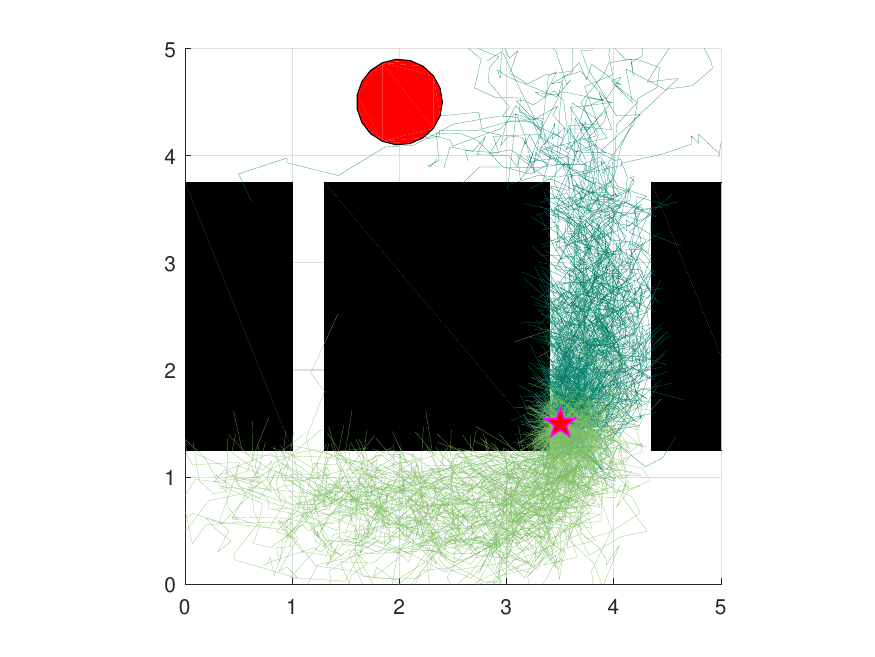}}
	\caption{Sample trajectories from (a) feedback policy and (b) open-loop control sequence for $b=0.5$.}
	\label{fig:c02}
\end{figure}

For the first example, we consider a simple two-dimensional stochastic single integrator in the environment having three obstacles that makes two paths with different width.
The dynamics and the input cost weight are given by:
$$\mathbf{f}(\mathbf{x}) = \mathbf{0},~G(\mathbf{x}) = I_2,~\text{ and}~R(\mathbf{x}) = 2I_2,$$
i.e., the position of a  robot in the configuration space, $\mathbf{x}\in \mathbb{R}^2$, is controlled by the velocity input, $\mathbf{u}\in\mathbb{R}^2$, while the objective of control is to reach the goal region while minimizing the cost function, $J = E\left[\phi(\mathbf{x}(t_f))+\int_0^{t_f}q(\mathbf{x})+\mathbf{u}^T\mathbf{u}dt\right]$.
The cost rate penalizes the collision with obstacles and the time length or the trajectory as:
\begin{equation}
q(\mathbf{x})= \left\{ \begin{array}{ll}
\infty & \text{if}~\mathbf{x} \in \mathcal{X}_{obs},\\
1 & \text{otherwise}.\end{array} \right.,
\end{equation}
and the final cost, $\phi$, encodes distance of a final state to the goal.

The state is driven also by a diffusion term that contains the 2-dimensional Brownian motion; two diffusion matrices are considered in this example for comparison:
$$
B(\mathbf{x}) = 0.1I_2,~0.3I_2.
$$
Through the path integral procedure, the time step for stochastic simulations and the number of samples for each reference trajectory are set as $\delta t = 0.1$ and $N = 300$, respectively.
The time horizon considered by the procedure is set as the minimum length of trajectories.
A feedback tracking controller is used for the importance sampler, $$g^{(h)}(t,\mathbf{x}) = \mathbf{u}^{(h)}_{ff}(t) + K(\mathbf{x}^{(h)}_{ref}-\mathbf{x}),~\forall h\in\{1,2\}$$  with $K = 1$, where $\mathbf{u}_{ff}(t)$ is the feedforward control input for $h$-th trajectory and the latter is trajectory stabilizer.
Finally, the period of receding horizon control is given by $\delta t$.

Fig. \ref{fig:c00} shows the results of the expansion phase (Algorithm \ref{alg:FMHT}) and \textsc{ExtractReference} function (Algorithm \ref{alg:ExtractReference}) for $\mathbf{x}_{cur} = [2.5, 0.5]^T$.
It is observed that two trajectories in different homology classes are returned, where only the obstacle in the middle makes a distinction between the trajectories, and they all connect the query state $\mathbf{x}_{cur}$ to the goal region.
When projecting the tree onto $H$-augmented space, the set of allowable $H$-signature value is defined as $\mathcal{A} \equiv \{z: |1-z_i|\leq H_{limit},~i = 2\}$\footnote{Note that only the $H$-signature for the middle obstacle is necessary in this example.} with $H_{limit}=0.6$ to extract trajectories in physically meaningful homology classes; otherwise, infinitely many trajectories that include paths revolving around the obstacle could be obtained.

Fig. \ref{fig:c01} depicts some snapshots of the receding-horizon control process in the execution phase (Algorithm \ref{alg:Execution})  with two different  diffusion matrices.
Note that, with large diffusion term, the effect of Brownian noise becomes so critical that the robot cannot pass through the narrow slit between the obstacles.
It is observed from the figure that when the noise is not critical, the robot goes to the goal region directly but it makes a detour when the noise increases.
It can be seen that, by considering topologically various trajectories as references, the path-integral formula provides comparative advantages between references.
\begin{table*}[h]\centering
	\caption{Success rate and path length from each algorithm}	\label{tbl:twoslit_comp}
	\begin{tabular}{rcccccc}
		\toprule
		\multirow{2}[3]{*}{Case ($b$)} & \multicolumn{2}{c}{Iterative-PI~\cite{theodorou2015nonlinear}} & \multicolumn{2}{c}{PI-RRT~\cite{arslan2014information}} & \multicolumn{2}{c}{PI-FMHT*} \\
		\cmidrule(lr){2-3} \cmidrule(r){4-5} \cmidrule(lr){6-7} 
		    & Success & Length            & Success & Length           & Success & Length         \\
		\midrule
		$0.05$ & 49 & 4.792  & 100 & 10.189	 & 100 & 4.535\\
		$0.1$  & 18 & 6.066  & 77  & 10.307  & 84 & 4.926\\
		$0.3$  & 3  & 25.887 & 13  & 11.641  & 41 & 11.120\\
		$0.5$  & 3  & 20.884 & 1   & 11.354  & 19 & 14.583\\
		\bottomrule
	\end{tabular}
\end{table*}

\begin{table}[h]\centering
	\caption{Success rate and path length from each algorithm without feedback}	\label{tbl:twoslit_comp2}
	\begin{tabular}{rcccc}
		\toprule
		\multirow{2}[3]{*}{Case ($b$)} & \multicolumn{2}{c}{PI-RRT~\cite{arslan2014information}} & \multicolumn{2}{c}{PI-FMHT*} \\
		\cmidrule(lr){2-3} \cmidrule(r){4-5} 
		& Success & Length            & Success & Length           \\
		\midrule
		$0.05$ & 100 & 10.546	& 100 & 4.583\\
		$0.1$  & 65  & 10.427	& 85 & 5.099\\
		$0.3$  & 13  & 11.528	& 38 & 13.340\\
		$0.5$  & 2   & 16.084 	& 8  & 20.190\\
		\bottomrule
	\end{tabular}
\end{table}
Compared to the existing methods that utilize an open-loop control sequence to guide the importance sampler (e.g.,~\cite{arslan2014information,ha2016topology}), sampling with the generalized feedback policy turns out to be more helpful in generating valuable trajectories.
For example, Fig. \ref{fig:c02} shows the sample trajectories obtained from the same trajectory sampler except $K=0$.
It is clearly shown that, in the sample trajectories, the states diverge from the reference as the simulations proceed.
These divergences, if they are too large, can prevent the importance sampler from utilizing valuable sample trajectories around the reference.
Beside the method adjusting the magnitude of noise as suggested in \cite{williams2015model}, this provides another degree of freedom to balance exploration and exploitation.

\subsection{Quadrotor Navigation in an Urban Environment}
\begin{figure}[t]
	\centering
	\includegraphics[width=1\columnwidth]{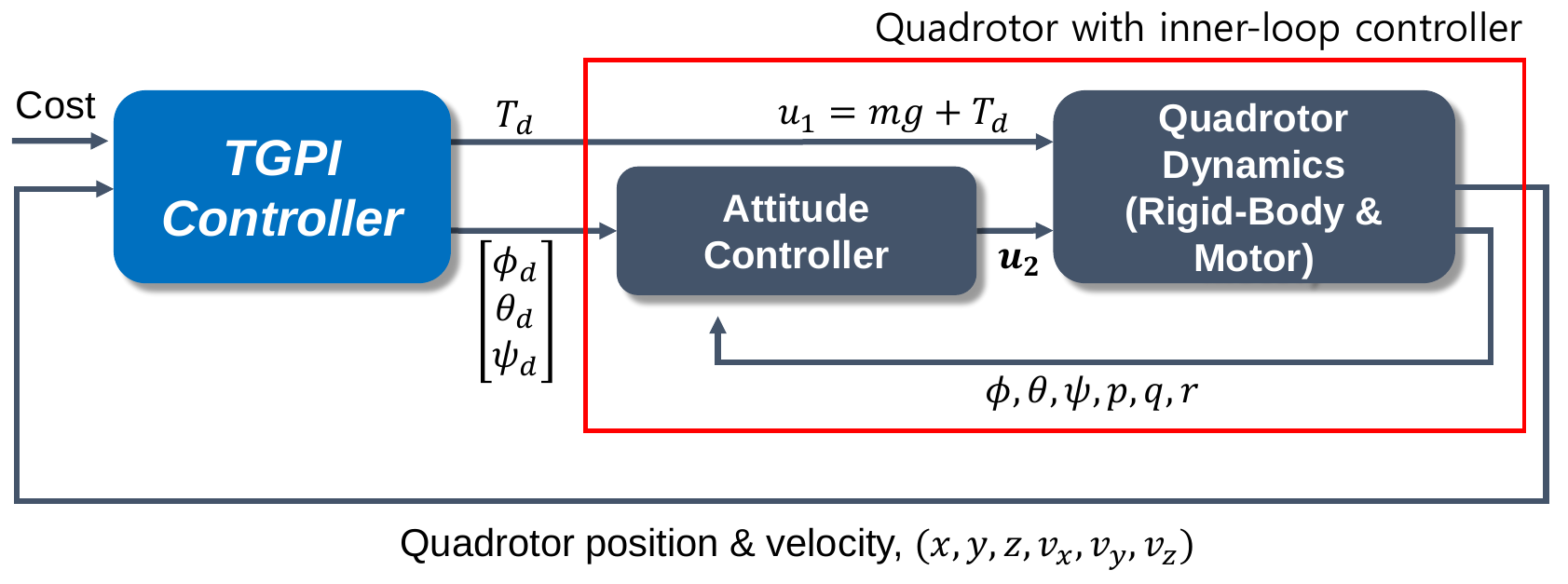}
	\caption{Quadrotor control scheme. The quadrotor is operated using the control input computed from the proposed Topology-Guided Path Integral (TGPI) Controller (Algorithm \ref{alg:Execution}).}
	\label{fig:quad_controlle}
\end{figure}
\begin{figure}[t]
	\centering
	\includegraphics[width=7cm]{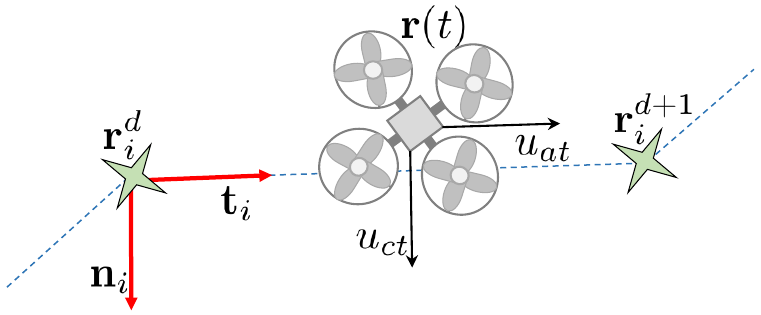}
	\caption{Path tracking scheme. The quadrotor is controlled to follow the path connecting the waypoints, and the control input is generated through the PD-controller.}
	\label{fig:path tracking scheme}
\end{figure}	
\begin{figure}[t]
\centering
\subfigure[]{
	\includegraphics*[width=.55\columnwidth,viewport= 0 0 380 225]{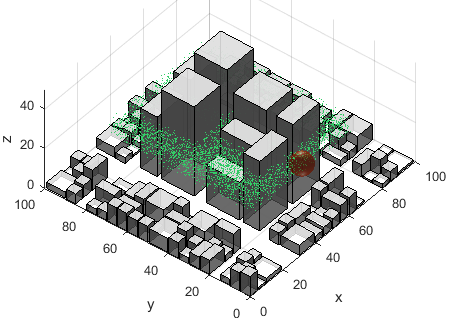}}
\subfigure[]{
	\includegraphics*[width=.4\columnwidth,viewport= 20 70 360 385]{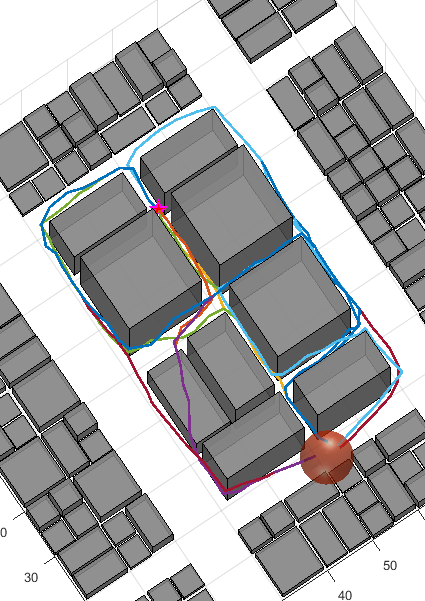}}	
\subfigure[]{
	\includegraphics*[width=.95\columnwidth]{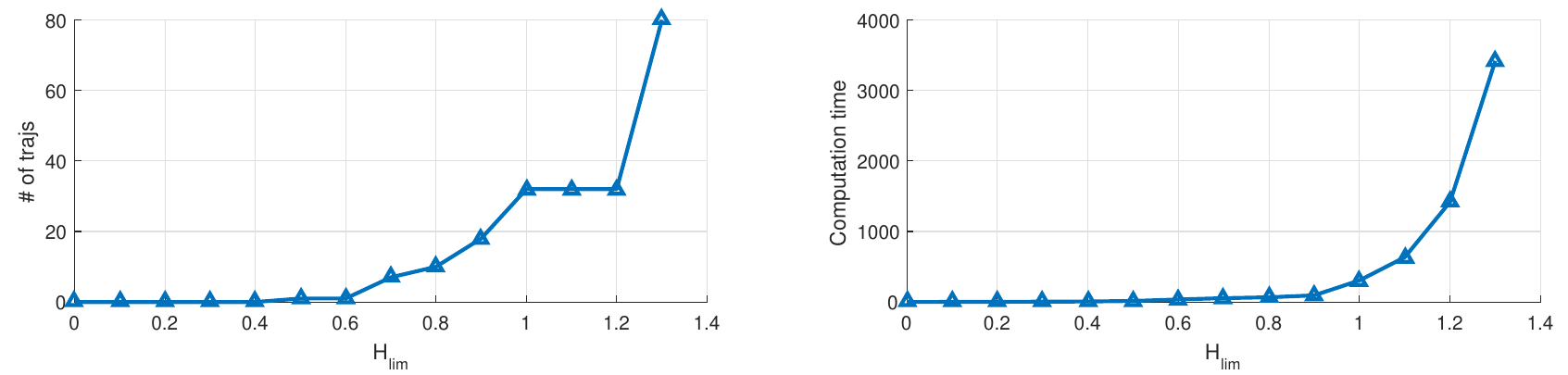}}
\caption{(a) The operation environment of the quadrotor in a complex urban environment. The red ball represents the final position where the quadrotor should reach, and the scattered green dots represent the sampled vertices taking the flight safety distance and operating altitude limitations of the quadrotor into account. (b) Results of topological path generation for quadrotor position $\mathbf{r}_{cur} = [50, 75, 25]^T$. (c) The number of trajectories w.r.t. $H_{lim}$}
\label{fig:quadTP}
\end{figure}	
\begin{figure*}[h!]
	\centering
	\subfigure[]{
		\includegraphics*[width=.38\columnwidth]{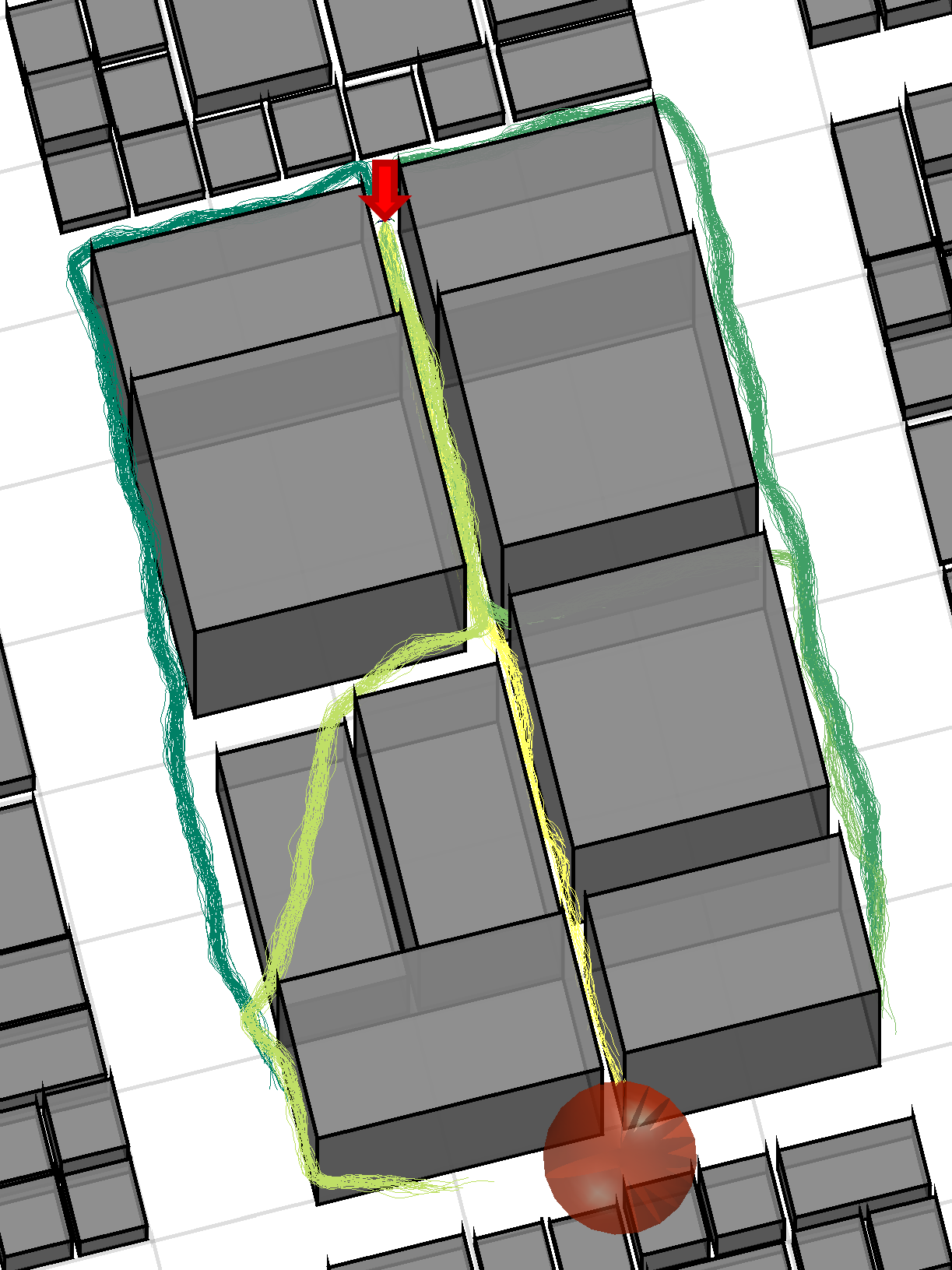}}
	\subfigure[]{
		\includegraphics*[width=.38\columnwidth]{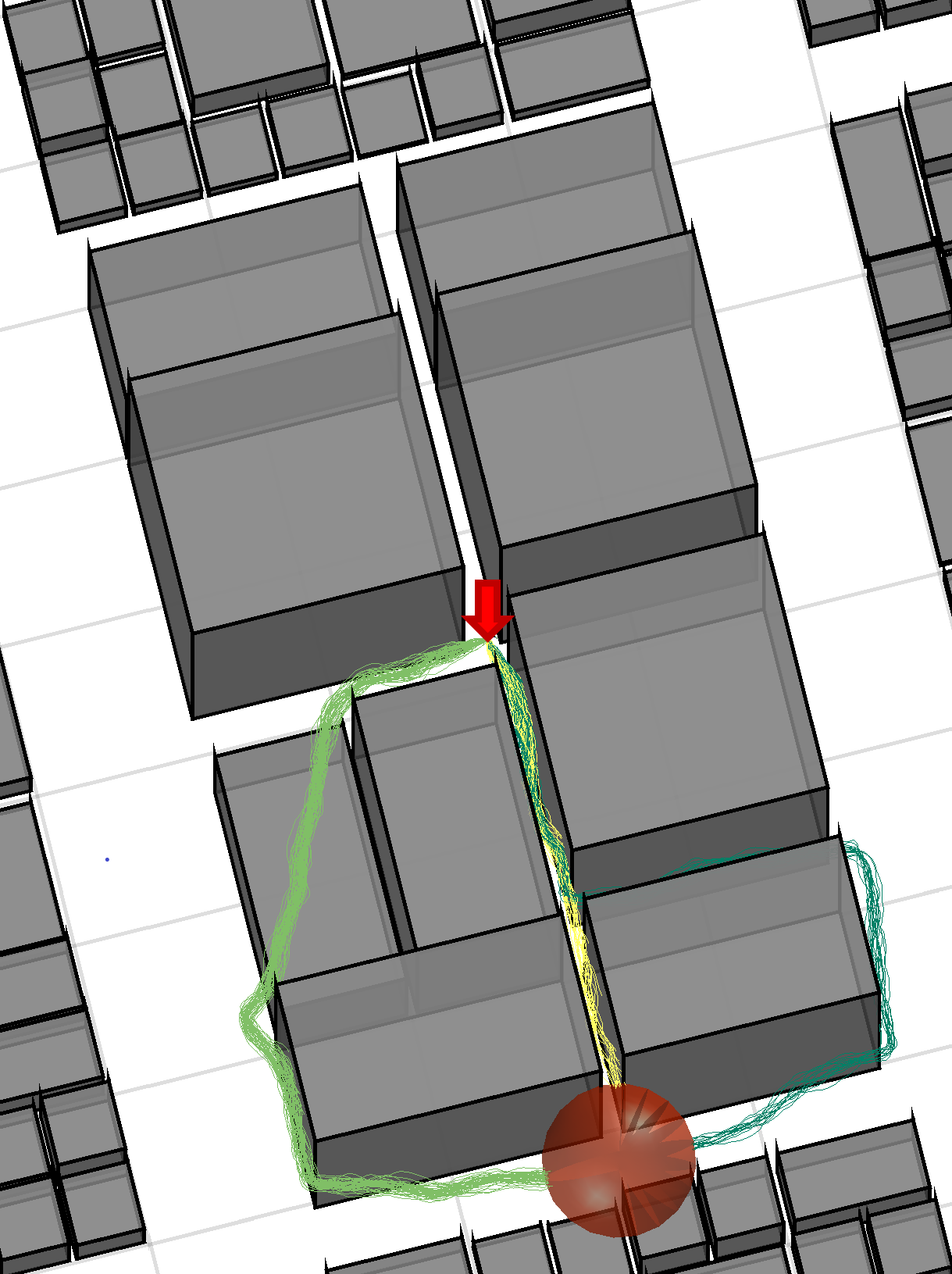}}
	\subfigure[]{
		\includegraphics*[width=.38\columnwidth]{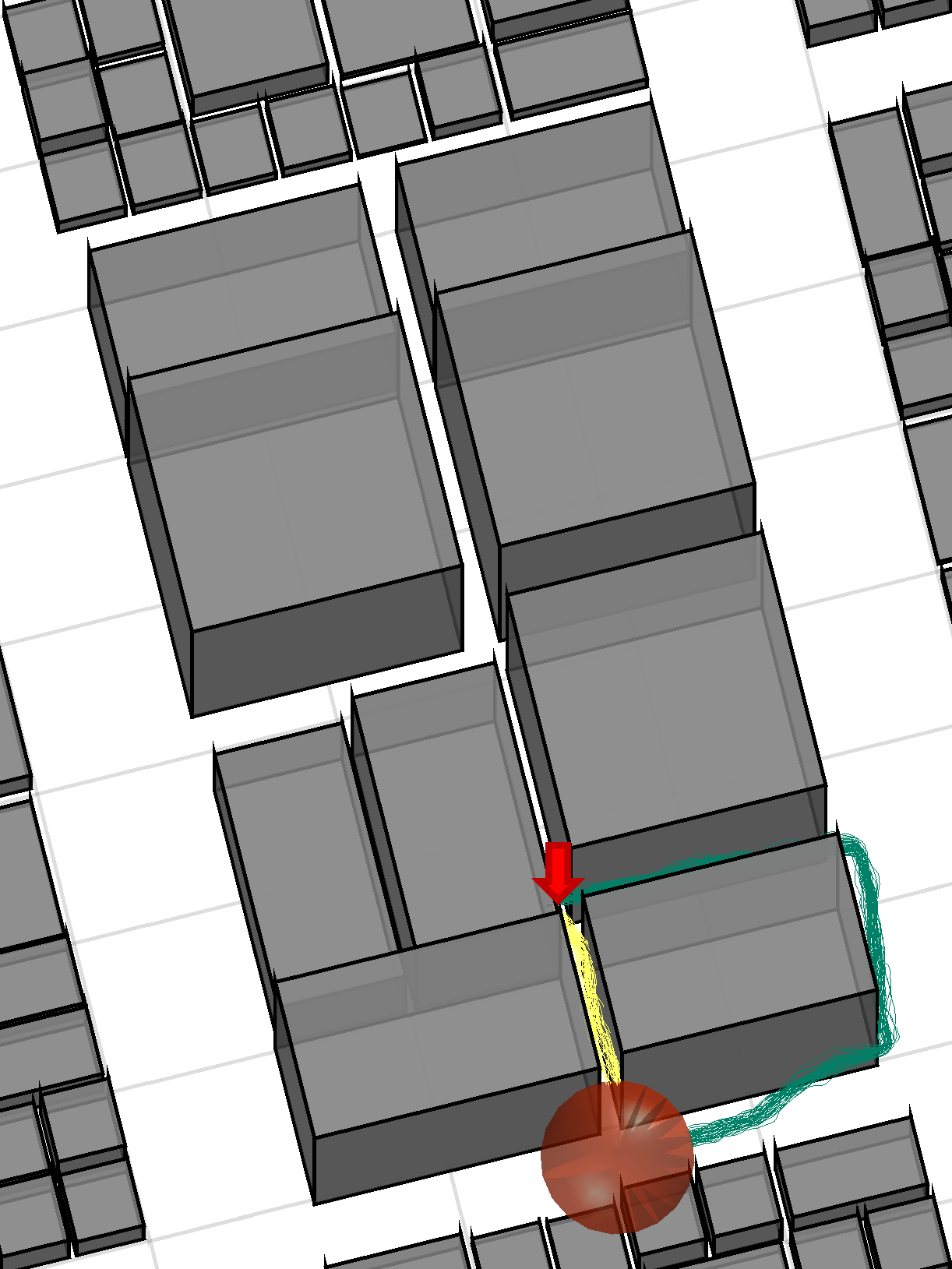}}
	\subfigure[]{
		\includegraphics*[width=.38\columnwidth]{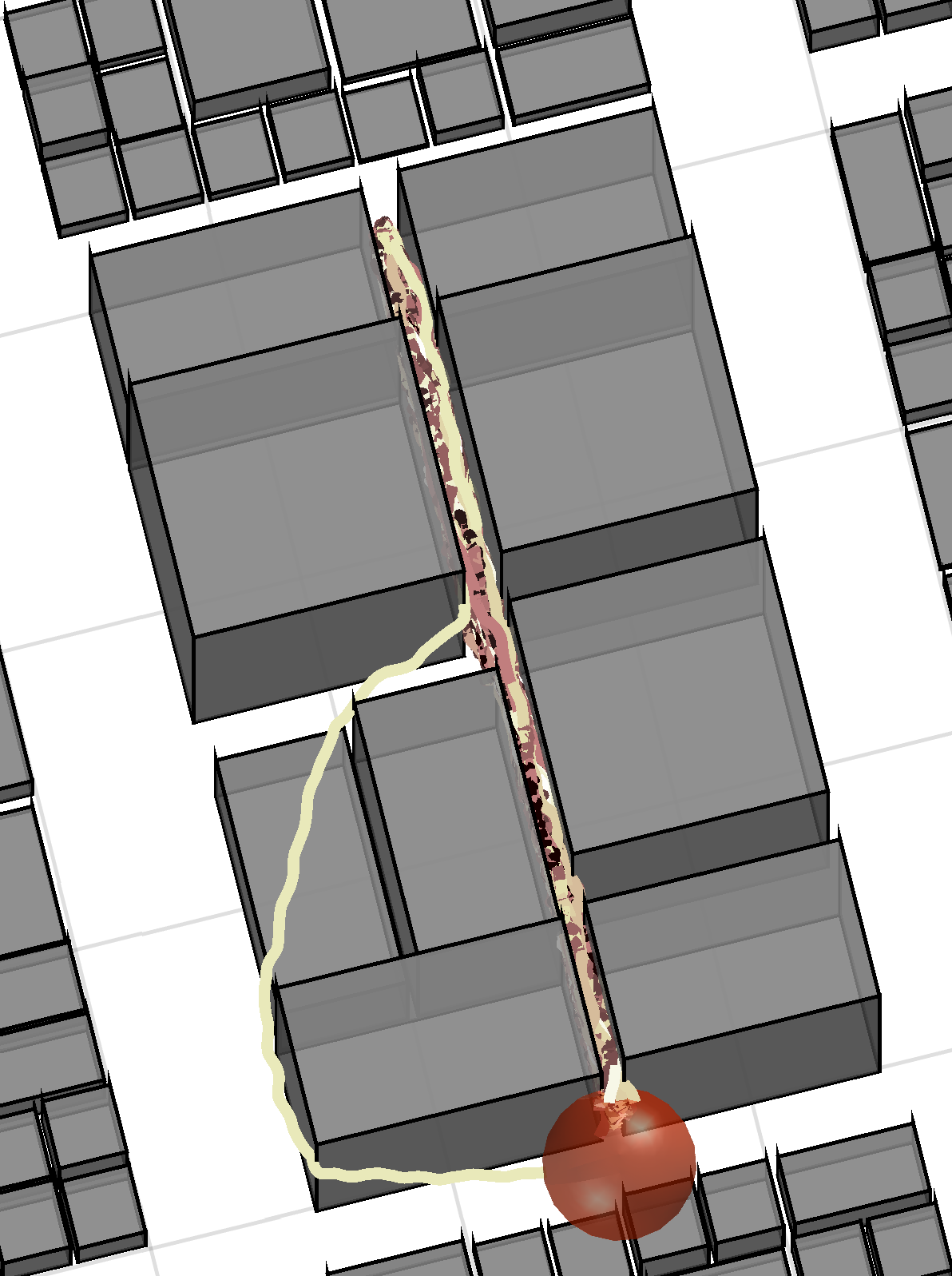}}

	\subfigure[]{
		\includegraphics*[width=.38\columnwidth]{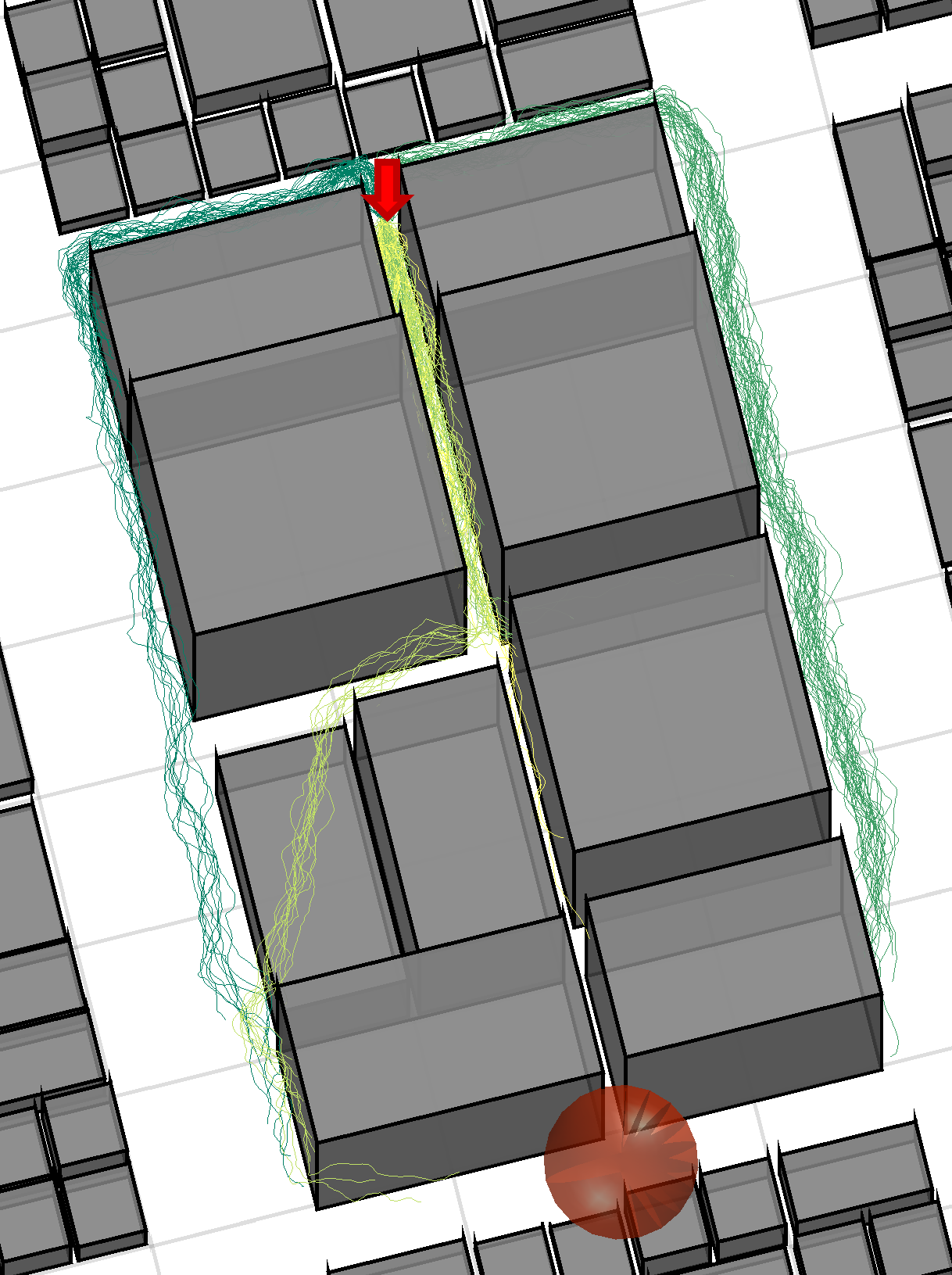}}
	\subfigure[]{
		\includegraphics*[width=.38\columnwidth]{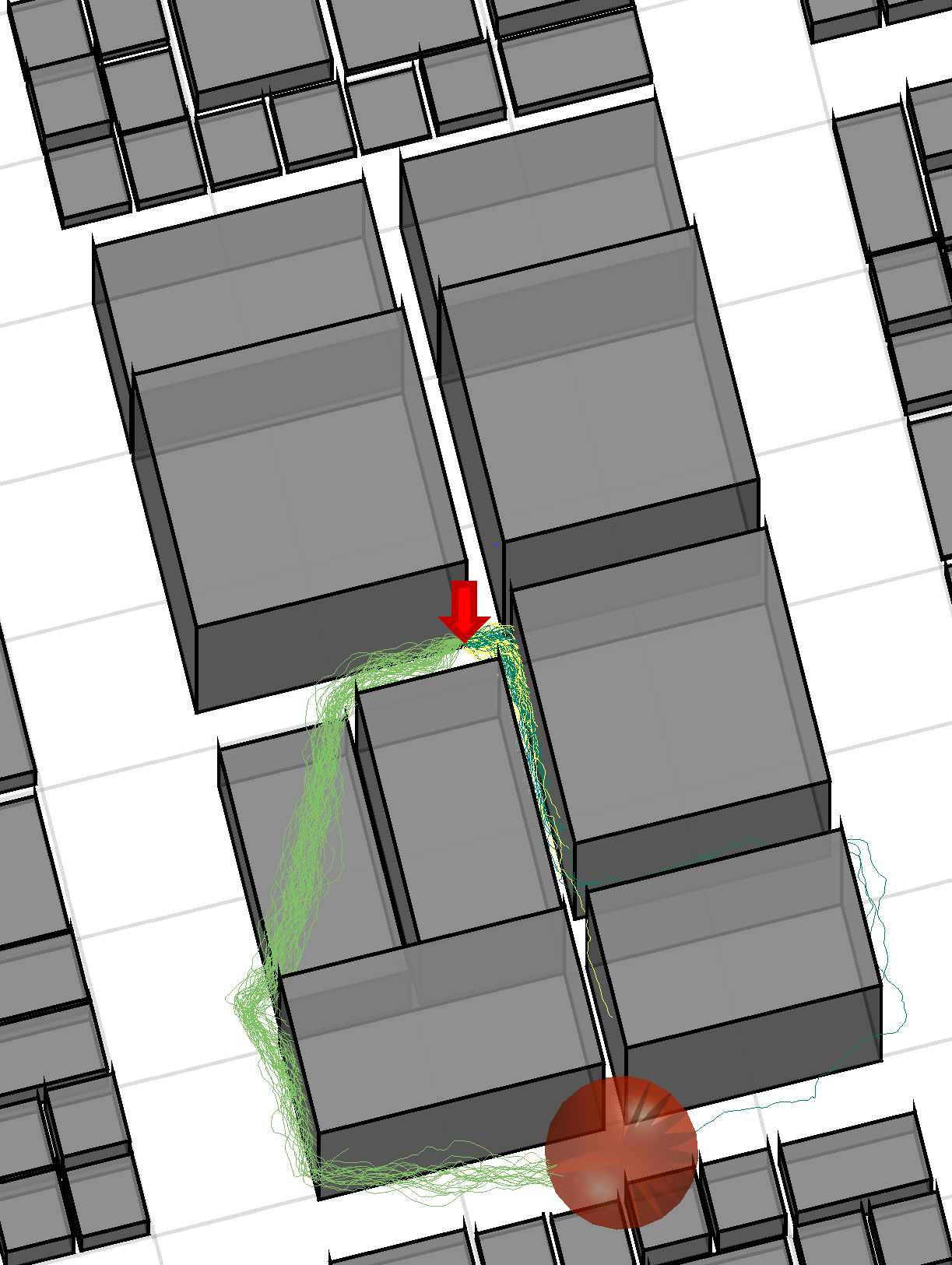}}
	\subfigure[]{
		\includegraphics*[width=.38\columnwidth]{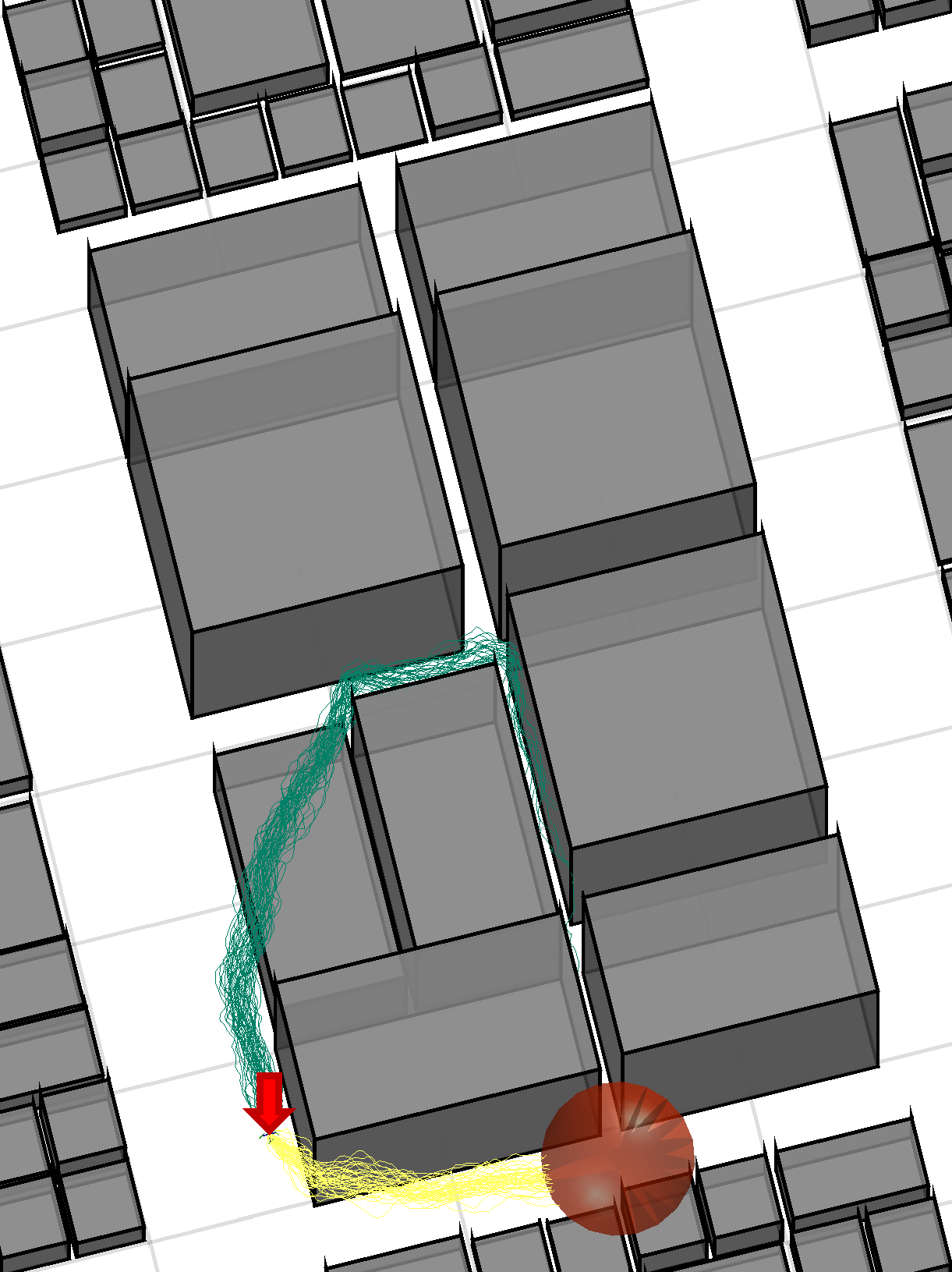}}
	\subfigure[]{
		\includegraphics*[width=.38\columnwidth]{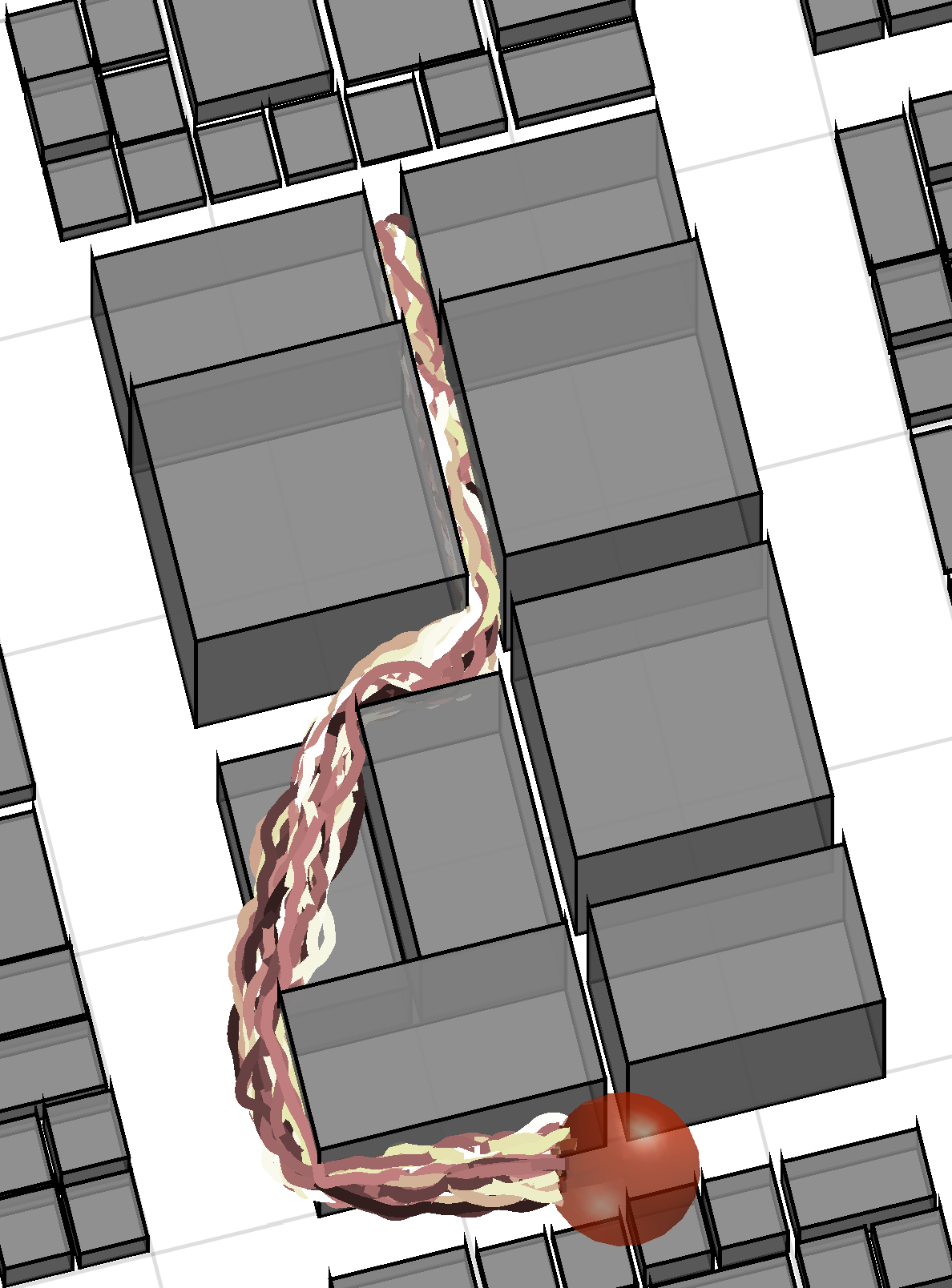}}

	\subfigure[]{
		\includegraphics*[width=.38\columnwidth]{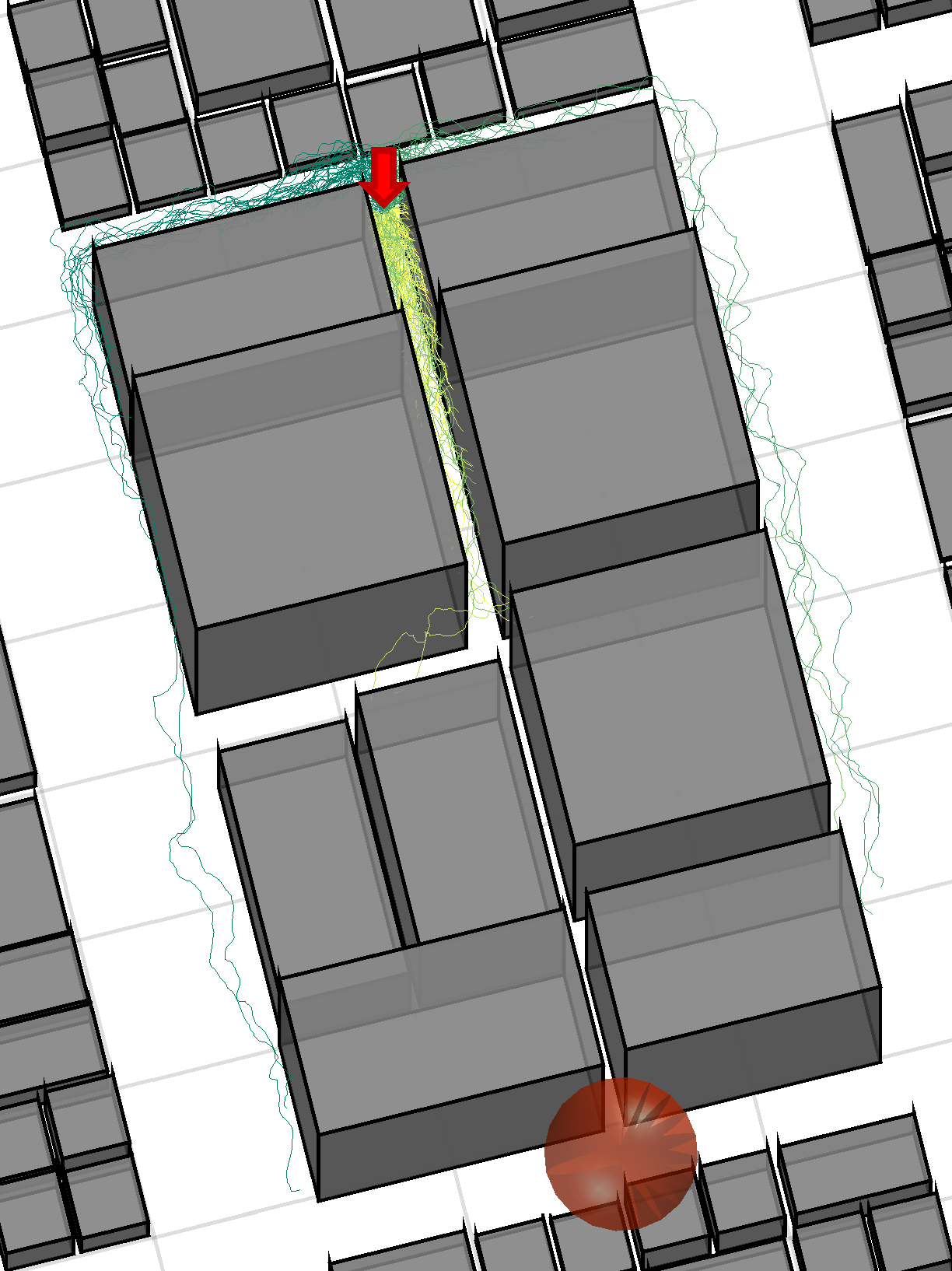}}
	\subfigure[]{
		\includegraphics*[width=.38\columnwidth]{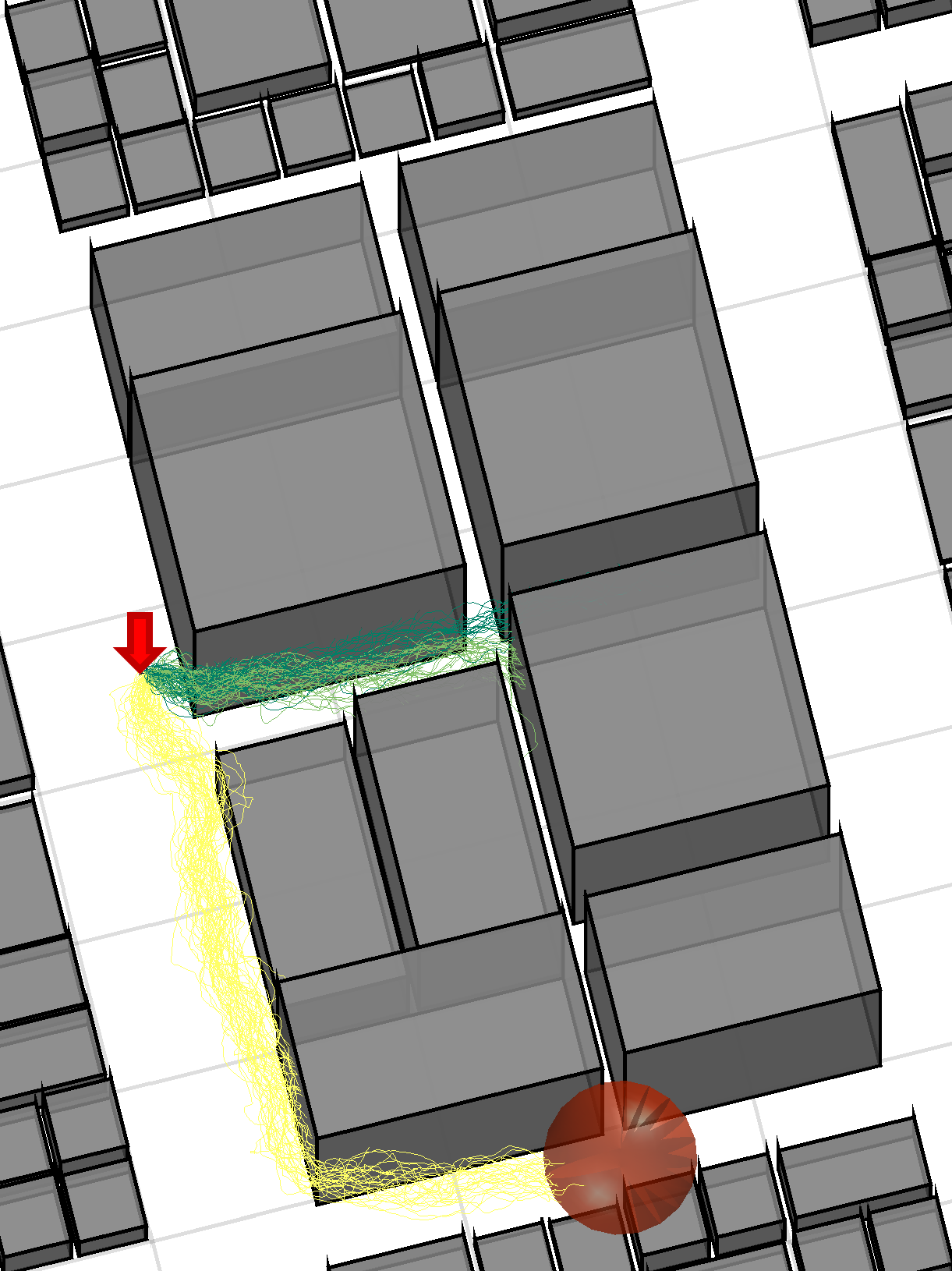}}
	\subfigure[]{
		\includegraphics*[width=.38\columnwidth]{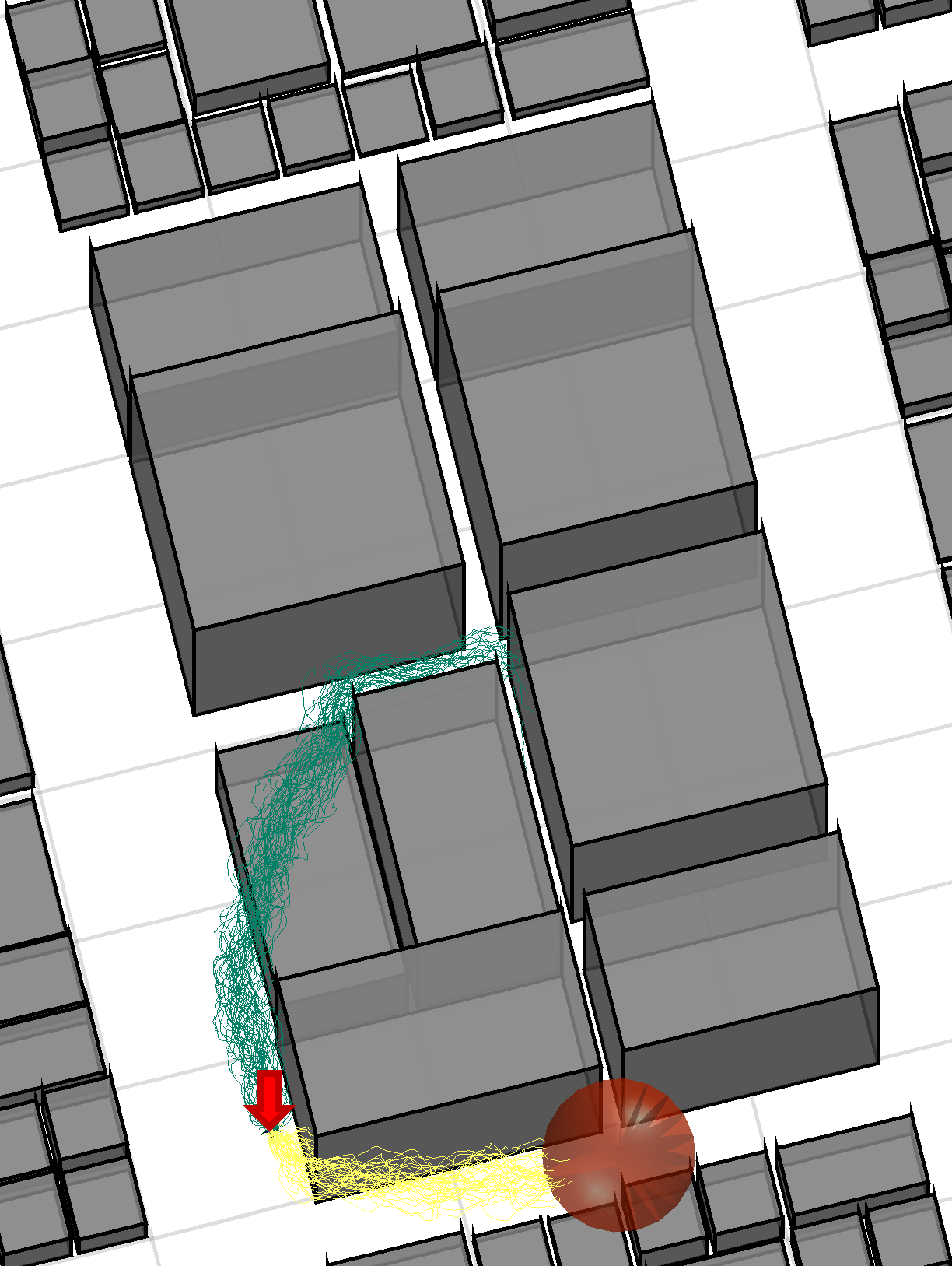}}		
	\subfigure[]{
		\includegraphics*[width=.38\columnwidth]{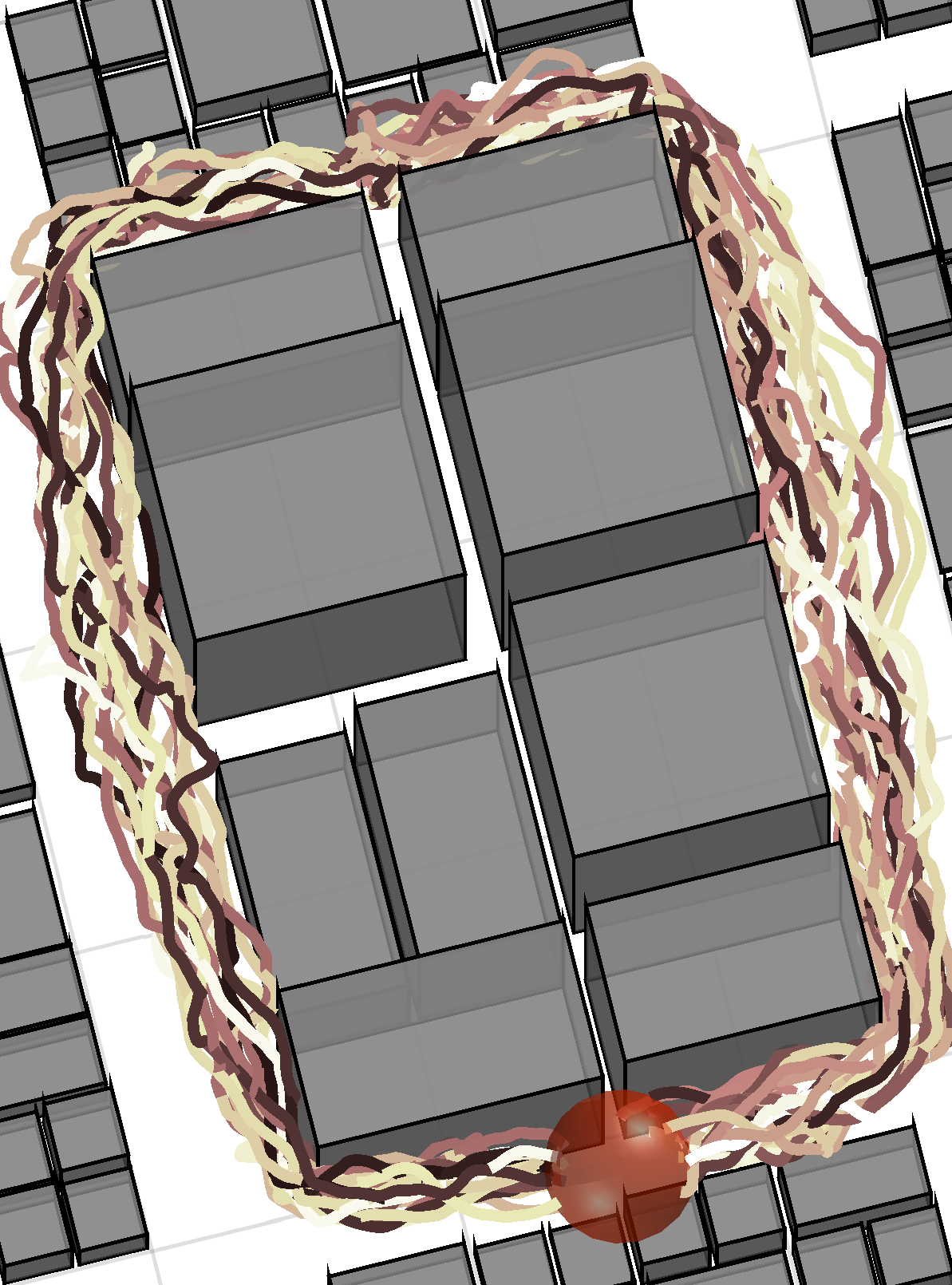}}
	\caption{Snapshots and resulting trajectories of the quadrotor navigation example, with (a)--(d) $\sigma = 1.5\times10^{-2}$, (e)--(h) $\sigma = 3\times10^{-2}$, and (i)--(l) $\sigma = 5\times10^{-2}$. A red arrow in the figure indicates the current position of the quadrotor.}
	\label{fig:quadTPGI}
\end{figure*}

The second example considers the situation in which a quadrotor is operated in a complex urban environment.
We used a 12-dimensional dynamic model for quadrotor control introduced in \cite{michael2010grasp}.
First of all, the full state of the quadrotor is given by the three-dimensional position $\mathbf{r}=[x,y,z]^T$, velocity $\mathbf{v}=[v_x,v_y,v_z]^T$, orientation $[\phi, \theta, \psi]^T$ (which represent roll, pitch, and yaw angles, respectively), and angular velocities $[p,q,r]^T$.
The inputs are given by the linear combinations of forces from each rotor, $F_i$, as:
\begin{align}
u_1 = \sum_{i=1}^{4}F_i,~\mathbf{u}_2 = L\begin{bmatrix}
0 & 1 & 0 & -1 \\ -1 & 0 & 1 & 0 \\ \mu_q & -\mu_q & \mu_q & -\mu_q
\end{bmatrix}\begin{bmatrix}
F_1\\F_2\\F_3\\F_4
\end{bmatrix},
\end{align}
where $L$ is the distance of the rotor axis from the center of the body, and $\mu_q$ is a coefficient for moment-force relation.
Then the 12-dimensional quadrotor dynamics is given by:
\begin{align}
&\dot{\mathbf{r}} = \mathbf{v},~\dot{\mathbf{v}} = \begin{bmatrix}0\\0\\-g\end{bmatrix} + \frac{1}{m}\begin{bmatrix}c\psi s\theta+c\theta s\phi s\psi\\s\psi s\theta-c\psi c\theta s\phi\\c\phi c\theta\end{bmatrix}u_1, \nonumber\\
&\begin{bmatrix}\dot{\phi}\\\dot{\theta}\\\dot{\psi}\end{bmatrix} = \begin{bmatrix}c\theta&0&-c\phi s\theta\\0&1&s\phi\\s\theta&0&c\phi c\theta\end{bmatrix}^{-1}\begin{bmatrix}p\\q\\r\end{bmatrix}, \nonumber\\
&\begin{bmatrix}\dot{p}\\\dot{q}\\\dot{r}\end{bmatrix} = -I^{-1}\begin{bmatrix}p\\q\\r\end{bmatrix}\times I\begin{bmatrix}p\\q\\r\end{bmatrix} + I^{-1}\mathbf{u}_2, \label{eq:dyn_quad_full}
\end{align}
where $g$, $m$, and $I$ denote the acceleration of gravity, the mass of the quadrotor and the moment of inertia matrix, respectively; Also, $c\cdot$ and $s\cdot$ are the cosine and sine functions, respectively.
Generally, the quadrotor embeds the PD-type attitude controller (shown as a red-box in Fig. \ref{fig:quad_controlle}) as:
\begin{equation}
\mathbf{u}_2 = I\begin{bmatrix}k_{p,\phi}(\phi_d - \phi) - k_{d,\phi}p \\ k_{p,\theta}(\theta_d - \theta) - k_{d,\theta}q \\ k_{p,\psi}(\psi_d - \psi) - k_{d,\psi}r\end{bmatrix},
\end{equation}
and its position is controlled by the thrust and desired orientation~\cite{michael2010grasp}.
We linearized the quadrotor dynamics at the hovering state (with $u_1=mg+T_d$ and the fixed yaw angle, $\psi=0$) and considered the linearization effect and the transient happened inside the red-box as a noise.
Our new control inputs, $\hat{\mathbf{u}} \in\mathbb{R}^3$, are then set to be proportionate to the desired pitch $\theta_d$, roll $\phi_d$, and thrust signal $T_d$ which are sent into the red box: $\hat{\mathbf{u}} =[g\theta_d,-g\phi_d,\frac{1}{m}T_d]^T$.
Also, the states are the position and velocity of the quadrotor (see Fig. \ref{fig:quad_controlle}).
Then, the reduced dynamics is given as:
\begin{align}
d\mathbf{r} = \mathbf{v}dt,~d\mathbf{v} \approx \hat{\mathbf{u}} dt+\sigma d\mathbf{w}, \label{eq:reducedD}
\end{align}
where $\mathbf{w}$ is a 3-dimensional Wiener process where the noise, $\sigma d\mathbf{w}$, can be considered as the difference between actual quadrotor dynamics and approximated dynamics or environmental factors influencing the motion of the quadrotor, e.g., wind, rain, snow, or other disturbances.

\begin{table}[h]\centering
	\caption{Success rate and path length}
	\begin{tabular}{ccccc}
		\toprule
		\multirow{2}[3]{*}{Case ($\sigma$)} & \multicolumn{2}{c}{PI-FMHT*} & \multicolumn{2}{c}{Tracking~\cite{hoffmann2008quadrotor}} \\
		\cmidrule(lr){2-3} \cmidrule(lr){4-5}
		& Success & Length & Success & Length \\
		\midrule
		$3\times10^{-2}$ 	& 100 & 63.4443 & 46 & 65.0434 \\
		$6\times10^{-2}$	& 57  & 96.7682 & 16 & 65.3812 \\
		$1\times10^{-1}$	& 53  & 126.4426 & 6 & 66.2251 \\
		\bottomrule
	\end{tabular}
\end{table}

In this example, we considered the path tracking controller, $\hat{\mathbf{u}} =g(t,\mathbf{x})$, when computing the sample trajectories described in (\ref{eq:dyn_control2}) and (\ref{eq:reducedD}).
We used a path tracking controller proposed in \cite{hoffmann2008quadrotor}.
The path tracking problem can be constructed by defining each path obtained from Algorithm \ref{alg:FMHT}
as a sequence of desired waypoints $(\mathbf{r}_{1}^{d},...,\mathbf{r}_{i}^{d},\mathbf{r}_{i+1}^{d},...)$ and defining the desired speeds of travel as $v_i^d$.
The geometry of the tracking problem is depicted in Fig. \ref{fig:path tracking scheme}.
Let $\mathbf{t}_i$ and $\mathbf{n}_i$ be a unit tangent vector of path connecting $\mathbf{r}_{i}^{d}$ to $\mathbf{r}_{i+1}^{d}$ and a unit normal vector of $\mathbf{t}_i$, respectively.
Then, given the current position of the quadrotor, the tracking errors consist of the cross track error $e_{ct}$ and the along track error $\dot{e}_{at}$ and are expressed as,
\begin{equation}
\begin{split}
e_{ct}(t) 		&= (\mathbf{r}_i^d-\mathbf{r}(t)) \cdot \mathbf{n}_i, \\
\dot{e}_{ct}(t) &= -\mathbf{v}(t) \cdot \mathbf{n}_i, \\
\dot{e}_{at}(t) &= v_i^d-\mathbf{v}(t) \cdot \mathbf{t}_i.
\end{split}
\end{equation}	
The control input of PD-controller is obtained using the tracking error,
\begin{equation}
\begin{split}
u_{at} &= K_{atp} \dot{e}_{at}, \\
u_{ct} &= K_{ctp} e_{ct}+K_{ctd} \dot{e}_{ct}, \\
\end{split}
\end{equation}
and the control input for tracking is then computed by adding the damping effect for stability:
\begin{equation}
\hat{\mathbf{u}}(t) = -K_{v}\mathbf{v}(t) + u_{at}(t)\mathbf{t}_i+u_{ct}(t)\mathbf{n}_i.
\end{equation}	
$\mathbf{u}_{PI}$ is computed from ...
Finally, the desired pitch, roll and thrust signal can be computed as follows:	
\begin{align}
\begin{bmatrix}
\theta_d(t) \\ \phi_d(t) \\T_d(t)
\end{bmatrix} =    \begin{bmatrix}
1/g & 0 & 0 \\ 0 & -1/g & 0 \\ 0 & 0 & m
\end{bmatrix} \mathbf{u}_{PI}.		\label{eq:quad_track}
\end{align}

The overall control scheme is shown in Fig. \ref{fig:quad_controlle}.
TGPI controller in the figure denotes the procedure in Algorithm \ref{alg:Execution};
it simulates stochastic dynamics \eqref{eq:reducedD} with the tracking controller \eqref{eq:quad_track} for a reference trajectory obtained by Algorithm \ref{alg:FMHT}.
The required speed of the quadrotor is 1m/s, the time interval for the stochastic simulation, $\delta t$, is set to 0.2 seconds and 30 sample trajectories are generated for each homology class.
Because the environment considered is too large, the time horizon of the stochastic simulatation is restricted to be less than 100 seconds.
In addition, as the previous example, the state cost rate, $q$, penalizes the collision with a building as $\infty$ and the final cost, $\phi$, encodes the shortest distance of the quadrotor to the destination at the end of the simulation.
Finally, the period of receding horizon control is given by $2\delta t$.	

Fig. \ref{fig:quadTP}(a) shows the operating environment of the quadrotor in this example.
To realize the actual situation, the state space is established by considering the safety distance between the buildings and the quadrotor, and by setting the limit of altitude.
The red ball and the scattered green dots represent the final position where the quadrotor should reach and the sampled vertices in Algorithm \ref{alg:FMHT}, respectively.
Fig. \ref{fig:quadTP}(b) depicts the reference trajectories obtained from Algorithm \ref{alg:FMHT}.
A lot of reference trajectories in different homology classes exist because of the environmental complexity, which causes the high non-convexity of the problem (with many local optima).

In this example, we performed simulations assuming three different levels of noise: $$\sigma = 1.5\times10^{-2},~3\times10^{-2},~\text{and}~5\times10^{-2}.$$
Fig. \ref{fig:quadTPGI} shows some snapshots of the simulation results at three different noise levels.
The higher noise results are placed in the lower rows. As can be seen from the results, the higher the noise level is, the more frequent the quadrotor collides with the building in the narrow passage, and in such a situation the quadrotor is controlled to detour the wide passage.
In summary, the proposed TGPI controller allows for the efficient computation of the optimal control that takes the level of noise into account while alleviating the issue of local optima.

\section{Conclusions}
This paper has addressed a class of continuous-time, continuous-space stochastic optimal control in the context of robot motion control in a complex environment. A path integral formula and an associated sampling method have been presented, and a motion planner, which embed topological information, has been developed to generate reference trajectories needed for the sampling procedure. An overall scheme has then been developed in a receding-horizon control framework. The proposed algorithm has been shown not only to provide a dynamically feasible and collision-free trajectory but also to effectively alleviate the undesirable convergence to local optima. Numerical examples have demonstrated the validity of the proposed approach.

\section*{Acknowledgment}
This work was supported by Agency for Defense Development (under contract \#UD150047JD).	


\bibliography{RAS_TGPI}
\end{document}